# Hierarchical Motion Consistency Constraint for Efficient Geometrical Verification in UAV Image Matching

**San Jiang [1], Wanshou Jiang [1,2,*]**

[1] State Key Laboratory of Information Engineering in Surveying, Mapping and Remote Sensing, Wuhan University, Wuhan, 430072, China; jiangsan870211@whu.edu.cn

[2] Collaborative Innovation Center of Geospatial Technology, Wuhan University, Wuhan, 430072, China

* Correspondence: jws@whu.edu.cn; Tel.: +86-27-6877-8092 (ext. 8321)

**Abstract:** This paper proposes a strategy for efficient geometrical verification in unmanned aerial vehicle (UAV) image matching. First, considering the complex transformation model between correspondence set in the image-space, feature points of initial candidate matches are projected onto an elevation plane in the object-space, with assistant of UAV flight control data and camera mounting angles. Spatial relationships are simplified as a 2D-translation in which a motion establishes the relation of two correspondence points. Second, a hierarchical motion consistency constraint, termed HMCC, is designed to eliminate outliers from initial candidate matches, which includes three major steps, namely the global direction consistency constraint, the local direction-change consistency constraint and the global length consistency constraint. To cope with scenarios with high outlier ratios, the HMCC is achieved by using a voting scheme. Finally, an efficient geometrical verification strategy is proposed by using the HMCC as a pre-processing step to increase inlier ratios before the consequent application of the basic RANSAC algorithm. The performance of the proposed strategy is verified through comprehensive comparison and analysis by using real UAV datasets captured with different photogrammetric systems. Experimental results demonstrate that the generated motions have noticeable separation ability, and the HMCC-RANSAC algorithm can efficiently eliminate outliers based on the motion consistency constraint, with a speedup ratio reaching to 6 for oblique UAV images. Even though the completeness sacrifice of approximately 7 percent of points is observed from image orientation tests, competitive orientation accuracy is achieved from all used datasets. For geometrical verification of both nadir and oblique UAV images, the proposed method can be a more efficient solution.

## 1. Introduction

Unmanned aerial vehicle (UAV) images have been extensively used in many applications, e.g., agricultural management (Habib et al., 2016), building model reconstruction (Aicardi et al., 2016) and transmission line inspection (Jiang et al., 2017), due to their high spatial resolution and flexible data acquisition. Usually, for most market-available UAV platforms, conventional navigation devices, namely, the combined GNSS/IMU (Global Navigation Satellite System / Inertial Measurement Unit) systems, cannot be used onboard for the direct geo-referencing (Turner et al., 2014), mainly because of the payload limitation of UAV platforms and the high costs of these devices. Therefore, prior to above-mentioned applications, image orientation is required to resume accurate camera poses for subsequent processing and interpretation, and reliable image matching guarantees the success and precision of image orientation.

In the literature, local feature based matching has become the dominant paradigm, which consists of three major steps: (1) feature extraction for individual images; (2) feature matching for image pairs; and (3) geometrical verification to eliminate outliers. Recent years have seen an explosion of activity in the areas of feature extraction and matching, which can be observed from the earliest corner detectors (Harris and Stephens, 1988) to the newly invariant detectors (Mikolajczyk and Schmid, 2005) in the fields of digital photogrammetry and computer vision. Invariant detectors describe the local regions of interest points with feature vectors, namely, descriptors of feature points, which simplifies feature matching by searching the nearest point with the smallest Euclidean distance between two descriptor sets. Although feature descriptors facilitate image matching with simple vector comparisons, initial candidate matches generated from the nearest-neighbor searching technique are inevitably contaminated by false matches, due to the only usage of local appearances for feature descriptor and similarity measurement. In addition, for obliquely captured images (Jiang and Jiang, 2017b), occlusions and perspective deformations would further cause a majority of false matches, even with the integration of the two well-known techniques for outlier elimination: cross-check (two features are the nearest neighbors of each other at the same time) and ratio-test (the ratio between the shortest and the second shortest distances is lower than a specified threshold) (Lu et al., 2016). Therefore, geometrical verification plays a crucial role in the pipeline of local feature based matching.

Geometrical verification is the step of computing a transformation from initial matches, and classifying them into inliers and outliers based on whether or not a point is geometrically consistent with the estimated transformation model, where inliers and outliers represent true matches and false matches, respectively. Generally, the strategies for geometrical verification can be categorized into two groups. For the first group, a geometrical transformation, e.g., a similarity or affine transformation, is explicitly estimated from initial candidate matches. Due to its ability to tolerate a large fraction of outliers, the random sample consensus (RANSAC) (Fischler and Bolles, 1981) method is one of the most popular tools for robust model estimation. The RANSAC algorithm operates in a hypothesize-and-verify framework in which subsets of input data points are randomly selected and model hypotheses are estimated from the subsets; each model is then scored using the entire data points, and the model with the best score is the solution. The RANSAC algorithm can provide a very accurate solution even with high levels of outlier ratios reaching to 70 percent (Chum and Matas, 2008). However, the computational costs of the basic RANSAC method increase exponentially with the percentage of outliers. To cope with this frequently arising problem, many variants of the RANSAC method have been designed and proposed to improve its efficiency in terms of hypothesis generation and model verification (Raguram et al., 2008), and promising results have been reported. In the work of Chum et al. (2003), a locally optimized RANSAC, namely, LO-RANSAC, was introduced to decrease the number of samples by integrating a local optimization stage into the classical RANSAC framework. Chum and Matas (2005) proposed a progressive sample consensus (PROSAC) for efficient hypothesis generation, which exploited the linear ordering structure of initial correspondences to draw samples, instead of the uniform sampling strategy used in the RANSAC method. An overview of recent research in the variants of the RANSAC method was comprehensively reviewed in Raguram et al. (2013).

In contrast to the explicit model estimation, methods from the second group implicitly find a geometrical transformation, which is estimated by using the Hough transform (HT) (Hough, 1962). The core idea of HT is that feature locations and transformation parameters are internally correlated; each feature location in the feature space votes for one bin or multiple bins in the parameter space, and peaks of bins reveal promising geometrical transformations between two images. Compared with RANSAC-based methods, the Hough voting scheme has two important advantages. First, it can tolerate higher outlier ratios based on the observation that transformations between false matches are not consistent, which leads to random votes in the voting space; consistent transformations between true matches result in centralized votes, which forms salient peaks in the voting space. Second, the Hough voting scheme replaces the hypothesize-and-verify framework of the RANSAC method with a direct voting strategy, and high efficiency can be achieved even for a degenerated configuration with a very high outlier ratio. Consequently, the Hough voting scheme has been reported in some research for the sake of efficient geometrical verification. In the work of Lowe (2004), the Hough transform that estimates a similarity transformation was utilized to find all feature clusters with at least three entries in a bin, and each cluster is then used in a geometrical verification procedure to refine the transformation. For image retrieval in large-scale scenarios, Li et al. (2015) introduced a strategy that uses pairwise geometric relations derived from the rotation and scaling relations between correspondences for match verification, and a reduced correspondence set was generated to accelerate geometrical verification, which was achieved by using the one-versus-one matching strategy and the Hough voting scheme. To reduce the time costs of hypothesis generation, Schönberger et al. (2016) integrated a Hough voting scheme into the traditional RANSAC method for the searching of the most promising transformations.

However, verification approaches based on the Hough voting scheme are not as accurate as the RANSAC-based methods, mainly because of either the coarse voting space quantization or the weak geometrical consistency, e.g., only the rotation and scaling parameters used for the transformation approximation between two images. Therefore, other attempts that incorporate spatial filters for outlier elimination have been exploited, which are used as a pre-processing or post-processing step before or after a RANSAC-based method. Usually, a spatial filter used as a pre-processing step aims to increase the ratio of inliers and accelerate the convergence speed of the RANSAC method. In the work of Sattler et al. (2009), a spatial consistency check was introduced to generate a reduced correspondence set with a significantly increased inlier ratio, which was implemented by taking into account the matching quality in a feature's spatial neighborhood. Considering image matching with multiple nearest neighbors for each feature point, Lu et al. (2016) proposed a straightforward and effective method to filter false matches from initial candidate matches using a geometrical consistency voting strategy. In the research, two transformation parameters, namely, the rotation and scale, were adopted to construct the voting space and to reduce the voting complexity, and promising results were presented when the inlier ratio was below 10 percent. On the contrary, a spatial filter used as a post-processing step intends to remove remaining false matches after geometrical verification, which is usually implemented by the analysis of local spatial relationships. Yao and Cham (2007) designed a motion consistent function to detect outliers based on the observation that true matches in a small neighborhood tend to have consistent location changes. Considering three local spatial relationship constraints, Hu et al. (2015) amended the standard feature matching pipeline to increase its reliability, rather than only using the appearance information. Compared with the former strategy, i.e., pre-processing, the post-processing strategy depends on a hypothesis that a global transformation has been established, and its performance is prone to be affected by outliers because only local relationships are used for spatial consistency check.

Consequently, the combination of the Hough voting scheme and the RANSAC algorithm can enhance both of their strengths to achieve high efficiency and high precision for geometrical verification. The Hough voting scheme is adopted to filter obvious outliers and increase inlier ratios of initial candidate matches; then, the RANSAC method is used to extract reliable inliers that are geometrically consistent with an estimated transformation. This combination has been reported in recent research, e.g., the work of Lu et al. (2016). However, as noted in Sattler et al. (2009), high computational costs can frequently arise due to pairwise geometrical comparisons, which is commonly used to derive the rotation and scale relations of two images. For high spatial-resolution images in the field of photogrammetry, a large number of initial candidate matches can be searched, especially for UAV images; besides, image matching with multiple nearest neighbors further increases the number of matches (Lu et al., 2016), which noticeably increases the complexity of the establishment of geometrical relations. Second, as observed from the above-mentioned research, almost all verification strategies are executed in the image-space, and a complex transformation with at least two parameters is mandatorily required, which leads to further increasing computational costs.

Considering different characteristics of images captured in the field of photogrammetry, some considerations for geometrical verification have been documented. For UAV image geo-registration, Zhuo et al. (2017) proposed a matching pipeline with pixel-distances as a global geometrical constraint, where a histogram voting technique for location changes of matches was used to verify initial matches after the elimination of differences in the rotation and scale. Tsai and Lin (2017) suggested checking whether the distances of feature points in horizontal and vertical directions are similar to others, because the UAV orthoimages have been coarsely aligned. Then, a simple "three-sigma rule" was utilized to eliminate false matches. These two techniques can achieve high efficiency with 1D-voting because complex transformations have been simplified as a 2D-translation, and pairwise geometrical comparisons for the estimation of rotation and scale parameters can also be avoided. Inspired by this observation, this paper exploits onboard GNSS/IMU data to achieve efficient geometrical verification in UAV image matching, which is implemented by simplifying the complex transformation and avoiding the pairwise geometrical comparison for parameter estimation. In our previous studies, relevant research using onboard GNSS/IMU data for match pair selection and geometrical rectification has been documented (Jiang and Jiang, 2017a, b).

This paper proposes an efficient strategy for geometrical verification in UAV image match. The basic idea is to simplify the complex geometrical transformation of correspondences in the image-space by their projection in the object-space. First, rough POS of each image is calculated by using on-board GNSS/IMU data, and feature points of candidate matches are projected onto a specified elevation plane. Second, two projected points of each initial candidate match form a motion in the object-space. Then, a hierarchical motion consistency constraint (HMCC) is designed and implemented for obvious outlier elimination. Finally, comprehensive analysis and comparison of the proposed algorithm are conducted by using real UAV images.

This paper is organized as follows. Section 2 details the HMCC, which is followed by the implementation of the HMCC-RANSAC algorithm in Section 3. The proposed scheme for geometrical verification is comprehensively compared and analyzed in Section 4, and some aspects are discussed in Section 5. Finally, Section 6 presents the conclusions.

## 2. Hierarchical motion consistency constraint

To achieve efficient geometrical verification in UAV image matching, this paper aims to exploit the use of onboard GNSS/IMU data for the simplification of the transformation model between correspondences. The workflow for UAV image matching is partitioned into two parts, as shown in Figure 1. The input images are two UAV images. For the first part labeled 1, feature points are extracted from individual images by using the SIFT detector (Lowe, 2004), and a 128-dimensional descriptor is assigned to one feature point based on local appearances around the point; feature matching operations are then conducted between two descriptor sets, which is achieved by searching the nearest feature point with the smallest Euclidian distance. Two well-known techniques, namely cross-check and ratio-test, are subsequently executed to detect false correspondences from initial matches. The first part represents a standard pipeline to obtain initial candidate matches, as documented in the work of Lowe (2004). Therefore, the primary contribution of this paper is presented in the second part of the workflow, which is designed as a geometrical verification step for false match elimination. For the improvement of efficiency, this procedure includes two steps: (1) the first-stage outlier elimination to increase inlier ratios, and (2) the second-stage outlier elimination with the RANSAC method to refine final matches. In this paper, the RANSAC method for the estimation of a fundamental matrix is utilized. Thus, the primary work of this paper is to implement the hierarchical motion consistency constraint (HMCC) algorithm for the first-stage outlier elimination, which is implemented as follows.

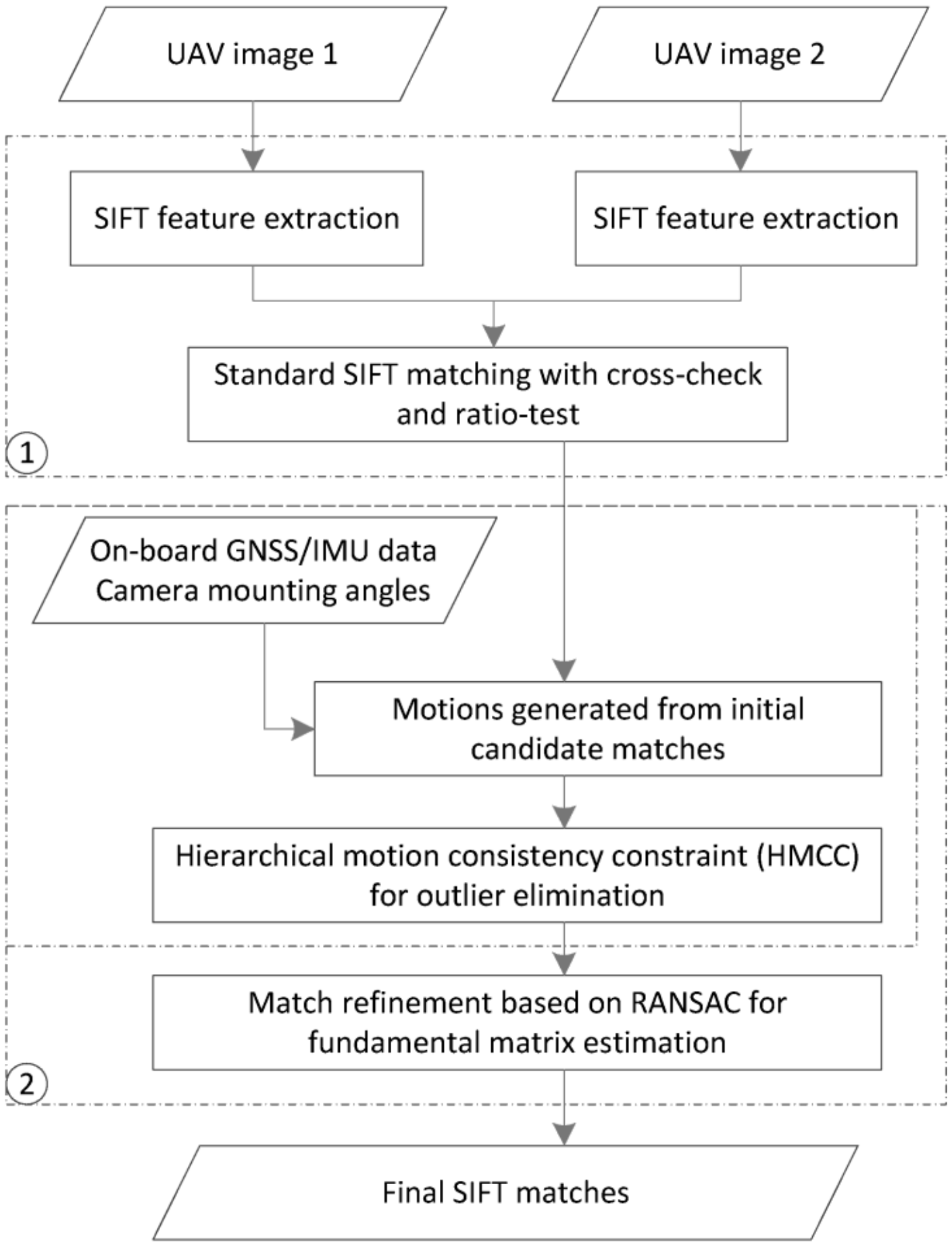

**Figure 1.** The overall workflow for UAV image matching. The work in the gray region indicates the primary contribution of this study.

*2.1. From initial candidate matches to motions*

The key idea of the HMCC algorithm is to convert the complex transformation model of correspondences in the image-space to a simple 2D-translation in the object-space. Therefore, prior to the projection transformation between the image-space and the object-space, camera poses should be calculated. For oblique UAV photogrammetry, onboard GNSS/IMU data provide the positions and orientations of the platform, and camera poses can be parameterized with the platform poses and their relative poses (Sun et al., 2016). Suppose that the relative pose between the platform and one camera is estimated and presented by a rotational matrix $r$ and a translational vector $t$; for one exposure station, the orientation and position of the platform are denoted by $R$ and $T$, respectively. Then, the actual camera pose can be calculated by Equation (1)

$$\begin{cases} R_c = rR \\ T_c = R^T t + T \end{cases} \tag{1}$$

where $R_c$ and $T_c$ are the rotational matrix and translational vector of the camera pose at the current exposure station, respectively. For nadir UAV photogrammetry, the above formulas are also applicable except that the relative rotation is represented by an identity matrix and the relative translation is indicated by a zero vector.

By using the calculated camera poses, two projection points of one correspondence can be computed by projecting feature points onto a specified elevation plane, which construct a directional vector that contains one starting vertex and one terminal vertex. In this study, the directional vector defines a primitive, termed motion, to implement the HMCC algorithm. The generation of motions from an image pair is implemented as follows: Given an image $I_i$, feature points are first extracted and denoted by $F(I_i)=\{x_j, y_j, d_j\}$ with feature location $p_j=(x_j, y_j)$ and feature descriptor $d_j$; for an image pair $(I_1, I_2)$, initial candidate matches $C=\{(f^1, f^2)\}$ are then established by matching two descriptor sets, where $f^1 \in F(I_1)$ and $f^2 \in F(I_2)$. With the aid of camera poses, feature locations $(p^1, p^2)$ of one candidate match $c$ in $C$ are projected onto a specified elevation plane $Z=Z_0$, as illustrated in Figure 2(a). The projection points of $p^1$ and $p^2$ are represented by two points $s$ and $t$ in the object-space, respectively. These two projection points form a motion $(s,t)$, where $s$ is the starting vertex and $t$ is the terminal vertex. To facilitate the analysis of motion consistency, a motion is defined as a directional vector with two properties, namely the length $L$ of the vector and the direction $\theta$ of the vector with respect to the X-axis of the object-space coordinate system, as shown in Figure 2(b). Thus, initial candidate matches of one image pair can generate a motion set, and the spatial relationships of correspondences in the image-space are converted to that in the object-space.

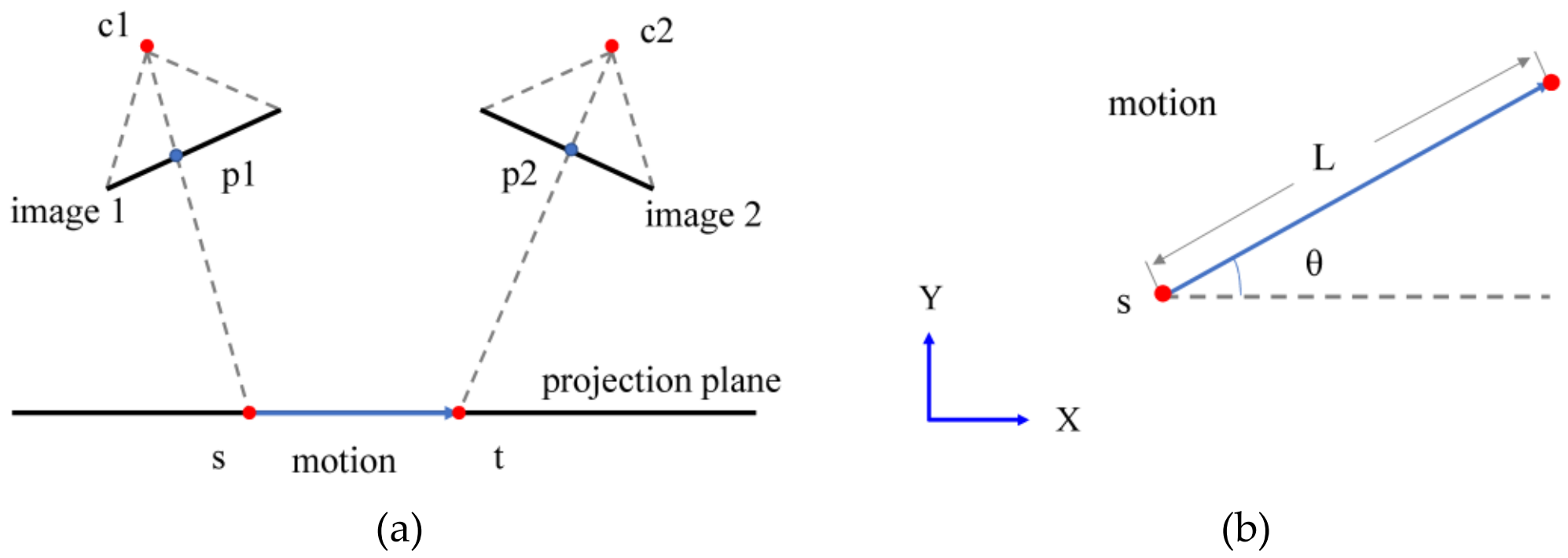

**Figure 2.** The generation of motions from initial matches: **(a)** the projection transformation from image-space to object-space; **(b)** the definition of motions as directional vectors.

After the projection transformation, spatial relationships of correspondences in the image-space are simplified as a simple 2D-translation in the object-space, as presented in Figure 2(b). Considering an ideal data acquisition with no terrain relief observed from fields and very precise camera poses are calculated, motions of true matches would degenerate to points when the height $Z$ of the projection plane is identical to that of the field. The degenerated motions are intersection points of corresponding rays (fired from projection centers and goes through corresponding points), which can be observed in Figure 2(a). However, in this study, camera poses are calculated from on-board GNSS/IMU data, which leads to the rough approximations of real poses. Thus, motions of true matches in the object-space cannot degenerate to points. In addition, for one image pair, bias errors from camera poses can be modeled as the relative orientation between two images, which causes the consistent rotation and translation of corresponding rays. For true matches, consistent motions with slight differences in length and direction are generated; on the contrary, for false matches, generated motions would not be consistent in terms of length and direction. This observation is the most important to design and implement the motion consistency constraint for outlier elimination.

*2.2. Hierarchical Motion Consistency Constraint - HMCC*

Spatial relationships of correspondences are reduced to a simple 2D-translation after the conversion from the image-space to the object-space. The 2D-translation of one correspondence in the object-space is defined as a motion, which is presented by a vector with two attributions, namely, the direction and length. Consistent motions with limited differences in direction and length are generated from true matches; on the contrary, inconsistent motions are created from false matches. This is the foundation for the motion consistency constraint. In this study, for efficient and reliable outlier elimination, a hierarchical strategy is adopted to implement the motion consistency constraint, namely HMCC, which consists of three major steps: (1) a global direction consistency constraint for the first-stage outlier elimination; (2) a local direction-change consistency constraint for the second-stage outlier elimination; and (3) a global length consistency constraint for the third-stage outlier elimination.

Suppose one image pair with $n$ initial candidate matches, the motions generated from initial candidate matches are denoted by $M = \{m_i : i = 1, 2, ..., n\}$; each motion $m_i$ consists of a starting point $s_i = (x_s, y_s)$ and a terminal point $t_i = (x_t, y_t)$; the direction of one motion is defined as the angle by which the X-axis of the object-space coordinate system can be rotated counter-clockwise to the corresponding vector of the motion; the length of the motion is defined as the norm of the corresponding vector. By using the starting point $s_i$ and the terminal point $t_i$, the direction and length of the motion $m_i$ can be respectively calculated by Equations (2) and (3)

$$direction = \arctan\left(\frac{y_t - y_s}{x_t - x_s}\right) \tag{2}$$

$$length = \sqrt{(x_t - x_s)^2 + (y_t - y_s)^2} \tag{3}$$

where $direction$ ranges from 0° to 360°, and $length$ is a real value. Based on the definition, the HMCC algorithm is implemented as follows.

For the first step of the HMCC algorithm, outliers are eliminated by using a global direction consistency constraint. This constraint is achieved based on the observation that directions of motions for true matches vary in a limited range; however, directions of motions for false matches would be random. Thus, a statistical analysis of directions is capable of detecting the majority of outliers in the first-stage of the HMCC algorithm. For one motion $m_i$, the direction $\theta_i$ is calculated by Equation (2); then, a direction list $dlist = \{\theta_i : i = 1, 2, ..., n\}$ can be obtained from $n$ motions in $M$, as shown in Figure 3(a), where motions are represented by directional vectors, and directions with respect to the X-axis of the object-space coordinate system are computed and listed at the bottom. Considering a possible high outlier ratio of initial candidate matches, a Hough voting scheme that is voted by the direction list $dlist$ is applied to remove outlier with random directions, because of its robustness to random noises (outliers). In Figure 3(a), the motion in orange is considered an outlier, and the others are inliers with consistent directions. The algorithmic implementation of the global direction consistency constraint is described in detail in Section 3.

The second step is to eliminate the remaining outliers. A local consistency constraint is designed based on direction-changes of motions, which is termed the local direction-change consistency constraint in this study. For one motion $m_i$, the $k$ nearest neighboring motions $M_{nn} = \{m_j : j = 1, 2, ..., k, j \neq i\}$ are searched from $M$; then, the direction-change between the motion $m_i$ and one of its neighbors $m_j$ in $M_{nn}$ is calculated by the absolute subtraction of the direction $\theta_j$ from the direction $\theta_i$, as presented by Equation (4)

$$dc_{ij} = \left| \theta_i - \theta_j \right| \tag{4}$$

where $|\bullet|$ denotes the operation of absolute subtraction. Therefore, a direction-change list $dclist_i = \{dc_{ij}\}$ between the current motion $m_i$ and its neighboring motions $M_{nn}$ is obtained. As shown in Figure 3(b), the red vector is the current motions, and its neighboring motions are rendered in green and orange; direction-changes of the current motion are sorted and listed at the bottom. In order to relief the influence of outliers, the direction-change of $m_i$ is defined as the median of the list $dclist_i$. Similarly, a Hough voting scheme that is voted by each motion's direction-changes is designed for the further elimination of outliers. In Figure 3(b), the motion in orange is considered an outlier. The algorithmic implementation of the local direction-change consistency constraint is described in detail in Section 3.

For the third step of the HMCC algorithm, motions with very short or long lengths are considered as outliers, because lengths of motions for true matches vary in a limited range; on the contrary, lengths of motions for false matches would be random. In this step, a global length consistency constraint is utilized to eliminate outliers that are in nearly the same directions of true matches. For one motion $m_i$, the length $l_i$ is calculated by Equation (3); then, a length list $llist = \{l_i : i = 1, 2, \ldots, n\}$ can be obtained from $n$ motions in $M$; finally, a simple z-score test is used to remove outliers, which is calculated by Equation (5)

$$zscore_i = \frac{l_i - \bar{l}}{\sigma} \tag{5}$$

where $\bar{l}$ and $\sigma$ are the average and the standard deviation of lengths in $llist$; $l_i$ is the length of the current motion $m_i$. In this study, a match with the z-score greater than three is considered as an outlier, which is the commonly used "three-sigma rule".

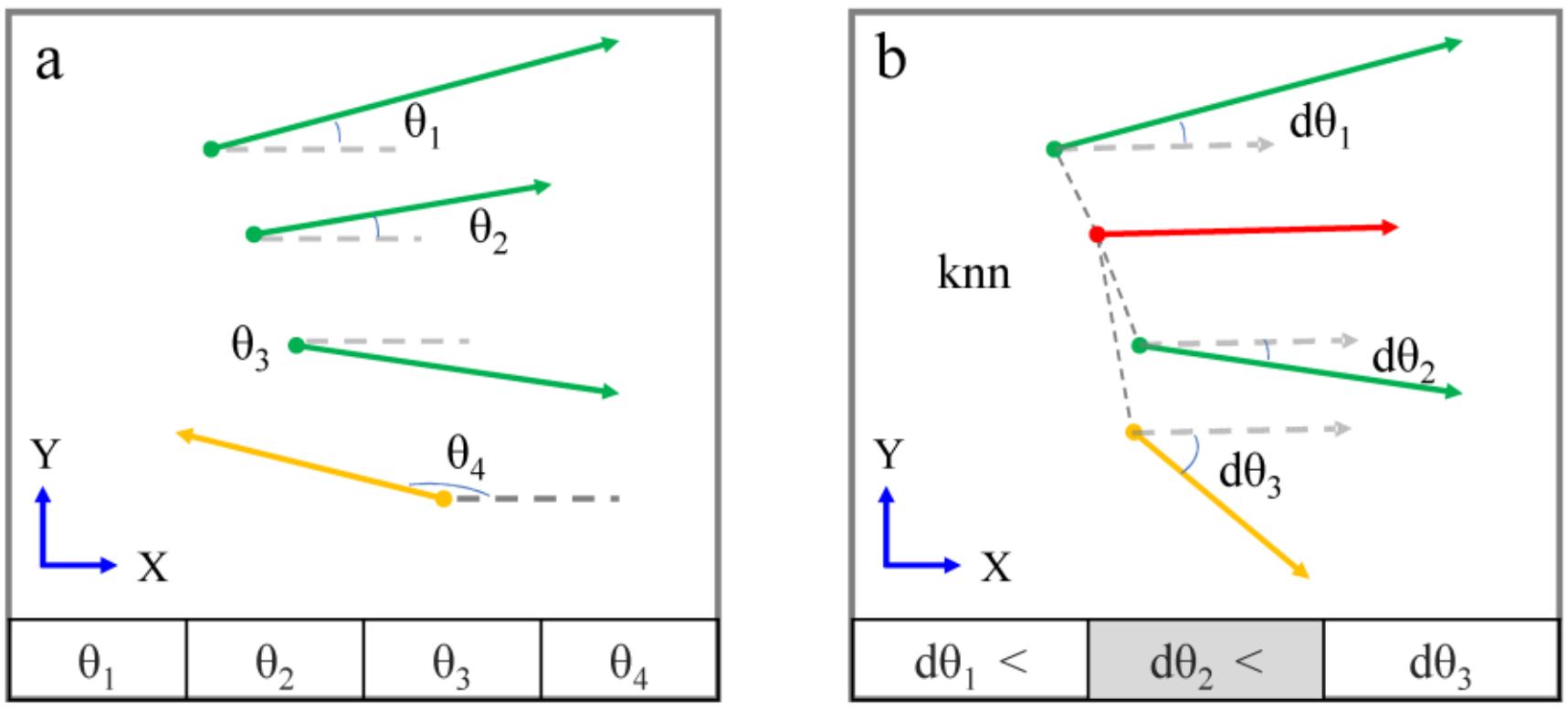

**Figure 3.** Illustration of the HMCC algorithm for outlier elimination: **(a)** the global direction consistency constraint; **(b)** the local direction-change consistency constraint. Motions are represented by directional vectors in the object-space. The motions corresponding to inliers and outliers are rendered in green and orange, respectively. The current motion is in red.

## 3. Implementation of the efficient geometrical verification

The workflow of efficient geometrical verification consists of the HMCC and RANSAC. The former aims to eliminate obvious outliers and increase the inlier ratio of initial candidate matches; the latter is used to refine final matches with a rigorous geometrical model. Due to its robustness to outliers, a Hough voting scheme is adopted in the first and second steps of the HMCC, which includes four main steps: (1) sampling of the voting space; (2) voting of the accumulation array; (3) peak determination; and (4) inlier extraction. Suppose $n$ motions $M = \{m_i : i = 1, 2, ..., n\}$ generated from initial candidate matches, the starting points of motions are extracted and denoted by $S = \{s_i : i = 1, 2, ..., n\}$; then, for one motion $m_i$, the $k$ neighboring motions $M_{nn} = \{m_j : j = 1, 2, ..., k, j \neq i\}$ can be searched from $M$, where the K-nearest neighbors algorithm (Cover and Hart, 1967) is indexed by the starting points $S$ and used for neighbor searching. The global direction consistency constraint and the local direction-change consistency constraint is implemented as follows.

**The voting scheme for the global direction consistency constraint**

(1) The variation range of the direction parameter $\theta$ is 0° to 360°, and the sampling interval of the voting space is set as 10°. Thus, a one-dimensional accumulation array $A$ with 36 bins is created and initialized to zero.

(2) For each motion $m_i$ in $M$, the bin index $idx_i$ calculated from its direction $\theta_i$ is recorded, and the votes of the bin with the index $idx_i$ in the accumulation array $A$ are increased by one. Thus, $n$ votes will occur in the accumulation array $A$.

(3) The peak bin with most votes is determined from the accumulation array $A$, and the neighboring bins within 5 steps away from the peak bin are also selected, as long as their votes are greater than 20 percent of the peak bin.

(4) For one motion $m_i$, it is considered as an inlier if its bin index $idx_i$ is within the index range of selected bins; otherwise, it is recorded as an outlier.

**The voting scheme for the local direction-change consistency constraint**

(1) The variation range of the direction-change parameter $\theta$ is 0° to 30°, and the sampling interval of the voting space is set as 3°. Thus, a one-dimensional accumulation array $A$ with 10 bins is created and initialized to zero.

(2) For one target motion $m_i$, $k$ neighboring motions $M_{nn} = \{m_j : j = 1, 2, ..., k, j \neq i\}$ are searched, and a direction-change item $dc_{ij}$ between the target motion $m_i$ and one of its neighboring motions $m_j$ in $M_{nn}$ is calculated, which results in a direction-change list $dclist_i = \{dc_{ij}\}$; then, for each item $dc_{ij}$ in $dclist_i$, its bin index $idx_{ij}$ is computed, and the votes of the bin with the index $idx_{ij}$ in the accumulation array $A$ are increased by one. In addition, the bin index $idx_i$ of the target motion $m_i$ is set as the median of the direction-change list $dclist_i$. Thus, $kn$ votes will occur in the accumulation array $A$.

(3) The peak bin with most votes is determined from the accumulation array $A$, and the neighboring bins within 3 steps away from the peak bin are also selected if their votes are greater than 40 percent of the peak bin.

(4) For one motion $m_i$, it is selected as an inlier if its bin index $idx_i$ is within the index range of selected bins; otherwise, it is considered as an outlier.

For neighbor searching, $k$ is set as 7 in this study. After the application of the HMCC algorithm, the inlier ratio of initial candidate matches is increased, which ensures the high convergent speed of the subsequent RANSAC method. In this study, the classical RANSAC method estimating a fundamental matrix with the seven-point algorithm (Zhang, 1998) is utilized to refine the final matches. To guarantee a high inlier ratio, the maximum residual error for the model estimation is set as 1.0 pixel in the RANSAC method. In conclusion, the above-mentioned procedures implement the efficient geometrical verification, namely HMCC-RANSAC, which is listed in Algorithm 1.

---

**Algorithm 1** Efficient Geometrical Verification (HMCC-RANSAC)

---

**Input:** $n$ initial candidate matches $C$, rough POS and projection plane $Z = Z_0$

**Output:** true matches $C_{fin}$

1: **procedure** HMCC-Filter

2: Generate motions $M = \{m_i : i = 1, 2, ..., n\}$ from initial candidate matches $C$

3: Vote with the global direction consistency constraint ( $M_d \leftarrow M$ )

4: Vote with the local direction-change consistency constraint ( $M_{dc} \leftarrow M_d$ )

5: Remove outlier with the global length consistency constraint ( $M_{red} \leftarrow M_{dc}$ )

6: Extract inliers according to the reduced motions ( $C_{red} \leftarrow M_{red}$ )

7: **end procedure**

1: **procedure** RANSAC-Estimation

2: Set the maximum inlier error $\varepsilon$=1.0

3: RANSAC for rigorous geometrical verification

4: Extract inliers according to the result of the RANSAC ( $C_{fin} \leftarrow C_{red}$ )

5: **end procedure**

---

## 4. Experimental results

In the experiments, a UAV dataset is used to evaluate the proposed algorithm. Image pairs with different configurations between oblique and nadir images are designed for the insight analysis of the HMCC algorithm. First, motions are generated from initial candidate matches. Second, the use of the three constraints, including the global direction consistency constraint, the local direction-change consistency constraint and the global length consistency constraint, is analyzed in the procedure of outlier elimination, which results in a reduced correspondence set with a significantly increased inlier ratio. Then, tests for its robustness to outliers are devised and conducted by using the reduced correspondence set. Finally, two methods that we refer to LO-RANSAC (Chum et al., 2003) and GC-RANSAC (Lu et al., 2016) are efficiently implemented, and comprehensive comparisons with the HMCC-RANSAC are performed in tests of feature matching and bundle adjustment (BA).

### *4.1. Dataset*

This dataset is collected from an urban area with the existence of dense buildings. A multi-rotors UAV platform equipped with one Sony ILCE-7R camera, dimensions of 7360 by 4912 pixels, is adopted for the acquisition campaign. For accurate photogrammetric measurement, the used camera has been calibrated using a calibration model with eight parameters: one for the focal length, two for the principal point, and three and two for coefficients of radial distortions and tangent distortions, respectively. Because the imaging system is equipped only one camera, two individual campaigns are conducted to obtain nadir (zero roll and pitch angles for camera installation) and oblique images (zero roll and 45° pitch angles for camera installation), respectively. The details of data acquisition can be found in our previous work (Jiang and Jiang, 2017b). At the flight height of 300 m with respect to the location from which the UAV takes off, a total number of 157 images are captured, and the GSD of nadir images is approximately 4.20 cm. The details of this data acquisition are presented in Table 1.

**Table 1.** Detailed information for data acquisition of the test site.

| Item Name | Value |
|---|---|
| UAV type | multi-rotor |
| Flight height (m) | 300 |
| Camera mode | Sony ILCE-7R |
| Sensor size (mm×mm) | 35.9 × 24.0 |
| Focal length (mm) | 35 |
| Camera mount angle (°) | nadir: 0 oblique: 45/-45 |
| Image size (pixel×pixel) | 7360×4912 |
| Number of images | 157 |
| Forward/side overlap (%) | 64/48 |
| GSD (cm) | 4.20 |

For system calibration, the lever-arm offset is assumed to be zero because the distance to the scene is much larger than the lever-arm offset. The bore-sight angles are much smaller than the camera installation angles. Thus, only camera mounting angles are used to approximate the relative poses of camera sensors with respect to UAV platforms. Image poses are calculated by using on-board GNSS/IMU data of the UAV platform and camera mounting angles in the acquisition campaign. First, GNSS/IMU data recorded in the navigation system are converted to a photogrammetric system. In this study, a local tangent plane (LTP) coordinate system with its origin located at the center of the test site is used. Second, the rough pose of each image can be calculated by using camera mounting angles, because they are the approximate respective displacements between UAV platforms and oblique cameras. The detailed process for image pose calculation is presented in Jiang and Jiang (2017b). By using the computed image poses, ground coverage of images for the dataset is illustrated in Figure 4, which are the projection of image corners on the average elevation plane of the test site.

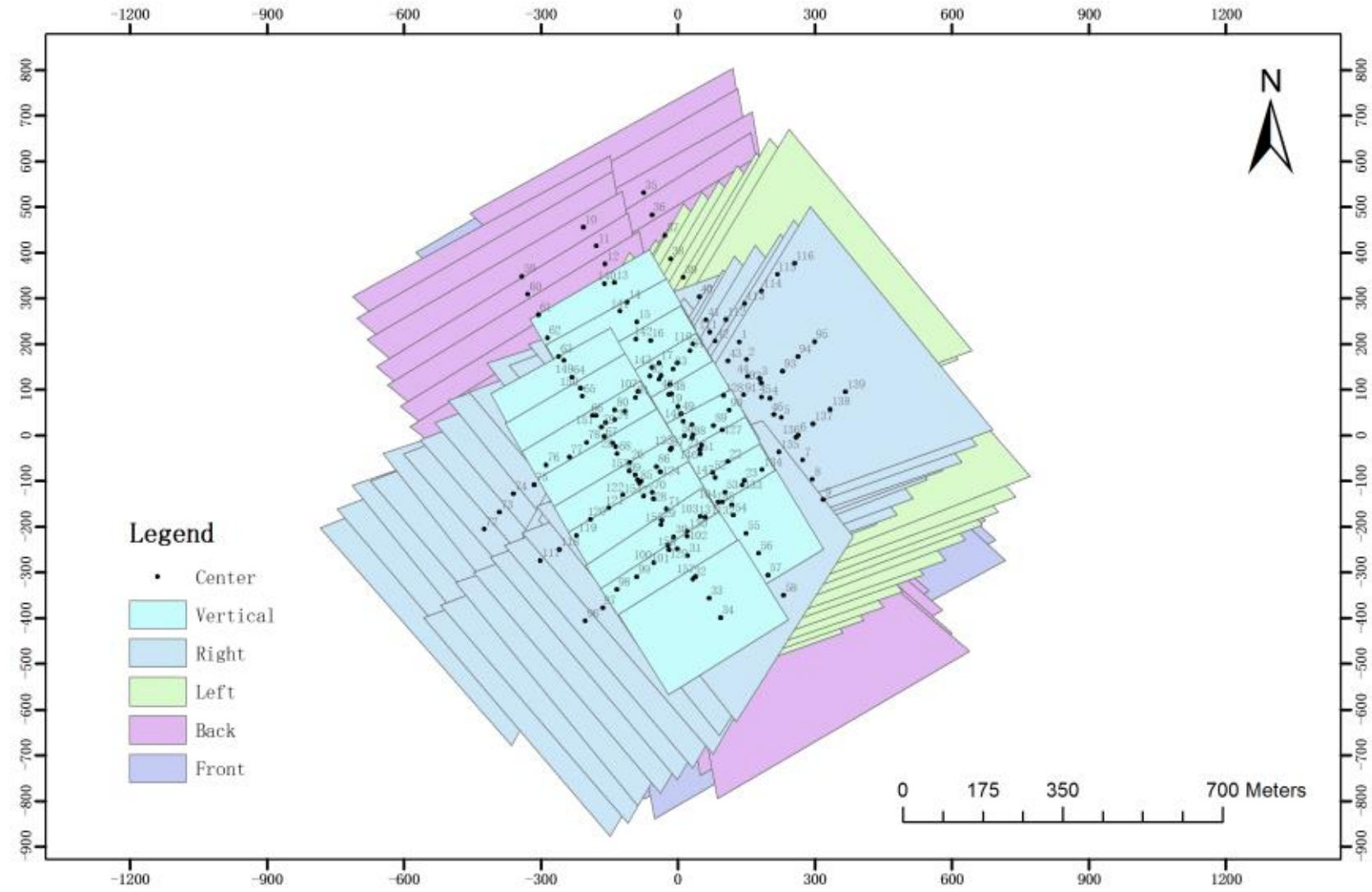

**Figure 4.** Ground coverage of the dataset (Jiang and Jiang, 2017b).

For performance evaluation of the HMCC algorithm on feature matching, four image pairs are configured using both nadir and oblique images in the dataset, as listed in Table 2. Images of the first pair come from the nadir acquisition campaign, and a large number of features can be matched, which is used to evaluate the performance of the motion consistency constraint on nadir images. The second to the fourth image pairs with increasing intersection angles contain oblique images, which aims to assess the influence of varying perspective deformations on the performance of the HMCC algorithm, as well as approving its adaption to oblique images. Due to large perspective deformations, many fewer initial candidate matches are extracted from the third and fourth image pairs. For the used dataset, direction indicators of the four image pairs are illustrated in Figure 5.

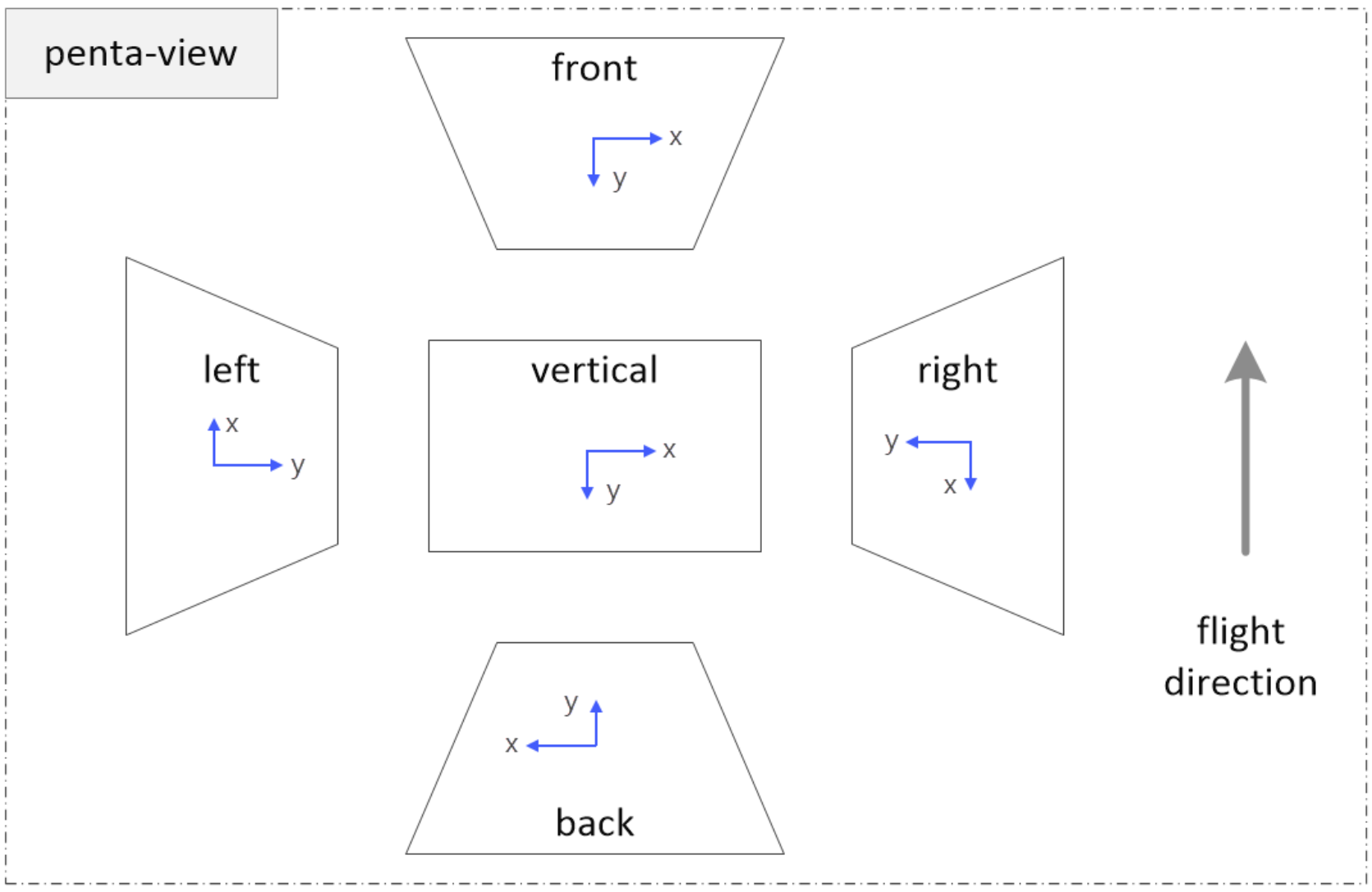

**Figure 5.** Illustration of a penta-view oblique photogrammetric system (Jiang and Jiang, 2017b). The acquisition of the used dataset is designed to simulate the ability of this system.

**Table 2.** Image pairs for performance evaluation of feature matching.

| No. | Direction | Description |
|---|---|---|
| 1 | V-V | Two nadir images |
| 2 | V-F | One nadir image and one front image |
| 3 | B-F | One back image and one front image |
| 4 | L-F | One left image and one front image |

*4.2 Motions generated from initial candidate matches*

The four image pairs listed in Table 2 are used to evaluate the performance of the HMCC algorithm on feature matching. The numbers of extracted feature points are 10,351 and 11,116 for the first image pair; 10,893 and 8,193 for the second image pair; 8,575 and 8,808 for the third image pair; and 8,348 and 8,808 for the fourth image pair. By searching the nearest feature point between two descriptor sets, initial candidate matches of each image pair are obtained, where two techniques, namely, cross-check and ratio-test, are adopted to reject outliers. In this study, the default parameters of the SIFT algorithm with ratio-test of 0.8 and max-distance of 0.7 are used in the feature matching stage. Initial candidate matches of the four image pairs are shown in Figure 6, where the numbers of the initial matches are 383, 134, 68 and 97 for the four image pairs, respectively. For a better interpretation of the matching results, the oblique images are geometrically transformed such that lines of true matches are approximately parallel to each other; however, lines of false matches intersect others. Observing the matching results, we can see that even with the two methods for outlier elimination, a large proportion of false matches exist in initial matches, especially for the last two image pairs, due to their relatively larger perspective deformations. By checking the matching results, inlier ratios of the four image pairs are approximately 52.5%, 32.8%, 26.5% and 27.8%, respectively.

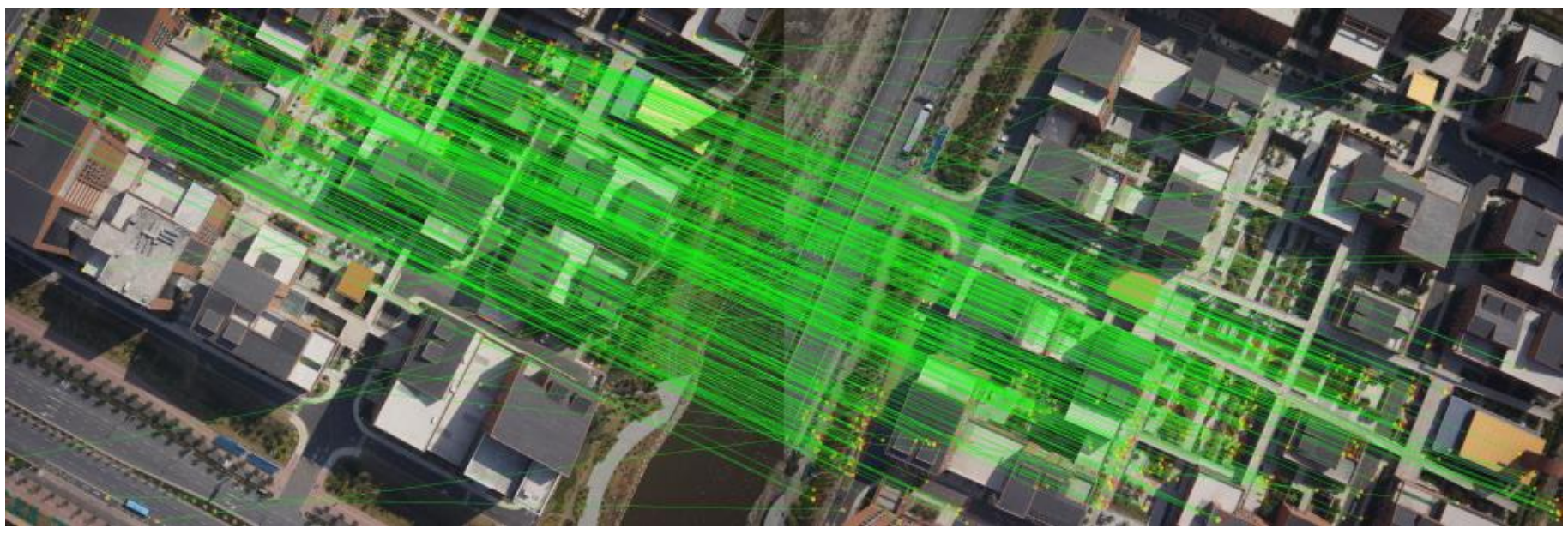

(a)

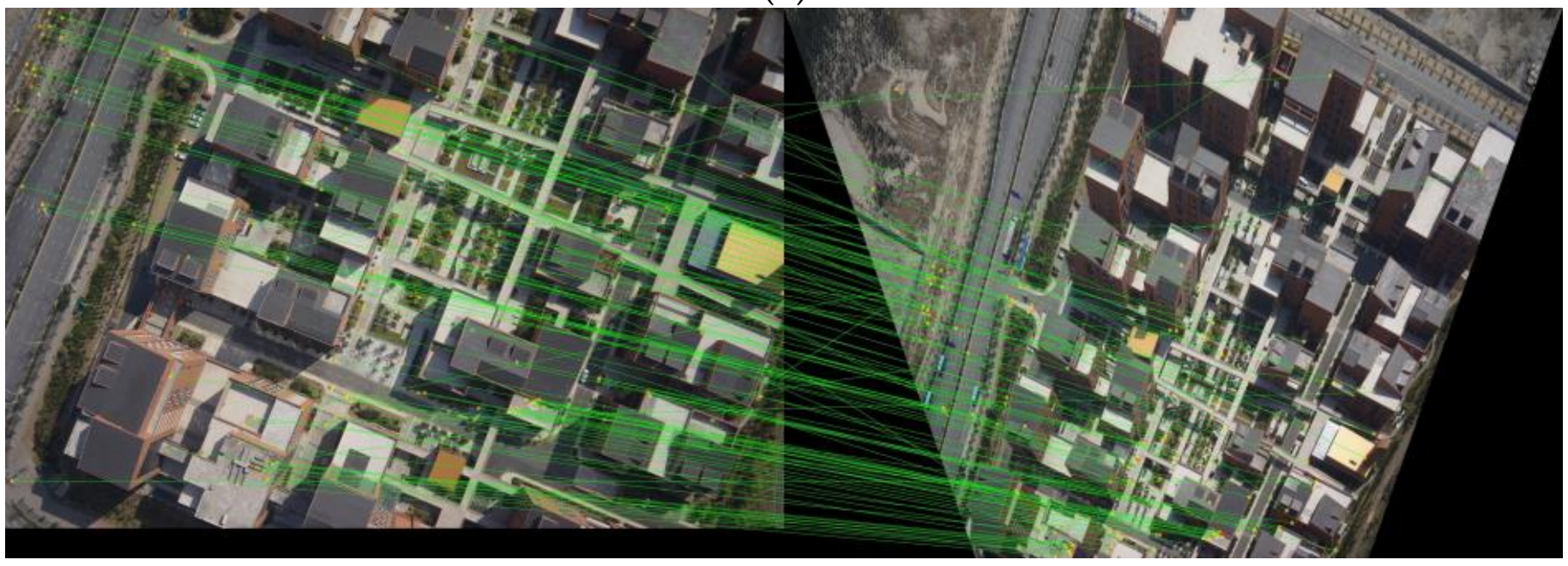

(b)

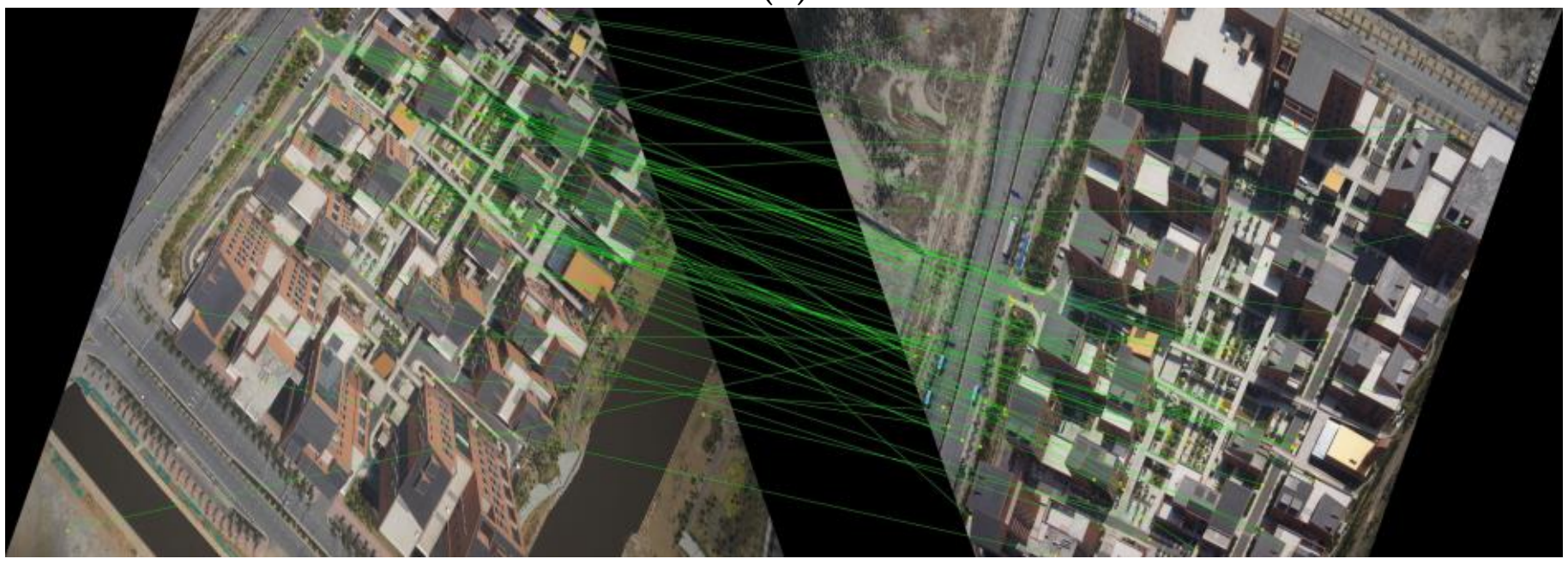

(c)

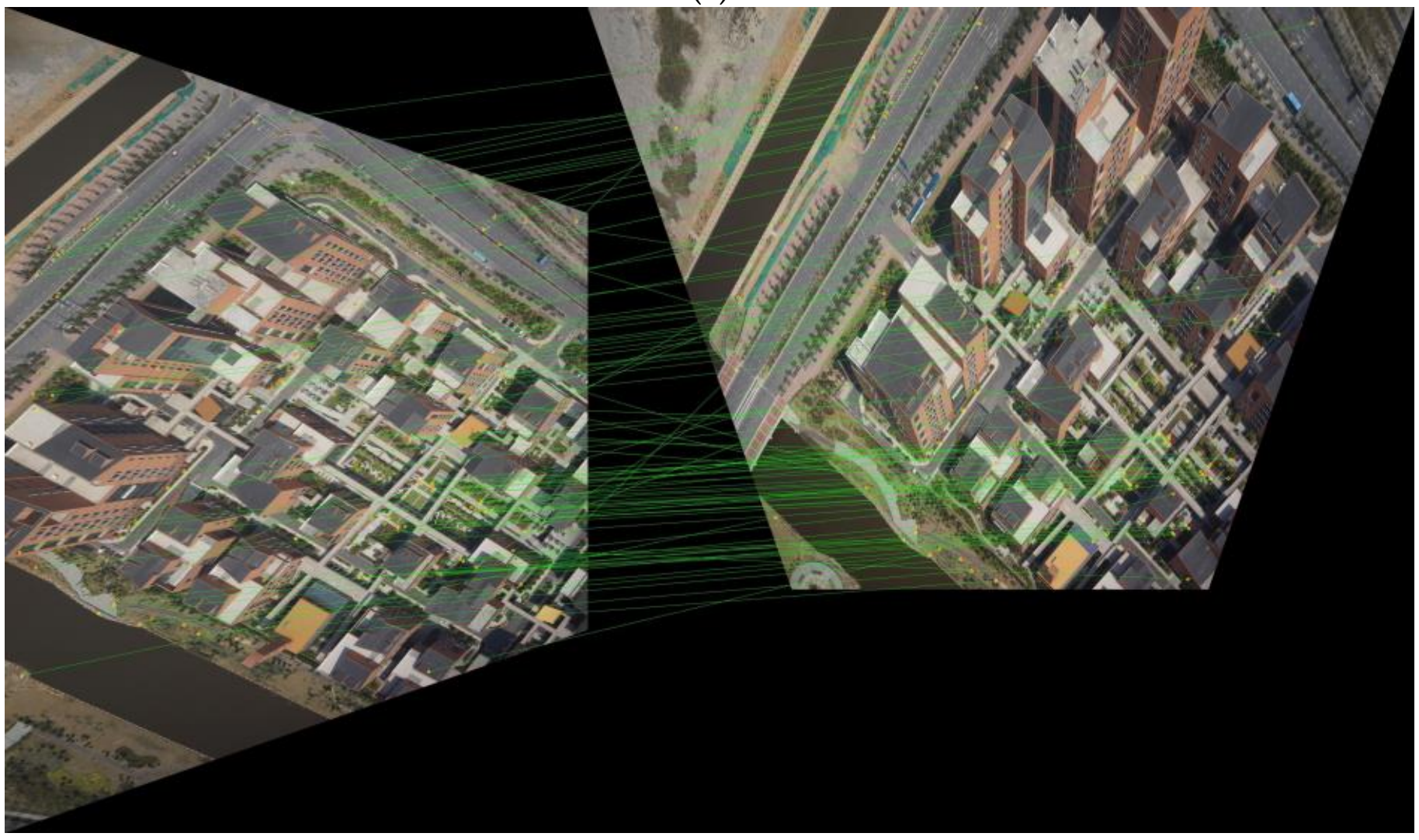

(d)

**Figure 6.** Initial candidate matches of the four image pair: **(a)** initial matches of the first image pair; (b) initial matches of the second image pair; (c) initial matches of the third image pair; (d) initial matches of the fourth image pair.

Prior to outlier elimination, motion generation from initial candidate matches is designed as the first step of the HMCC algorithm, which converts the complex relationship of matches in the image-space into a 2-dimensional translation in the object-space. Motions are generated by projecting the feature points of matches onto an elevation plane of the test site, as described in Section 2.1. In this study, the average altitude of the test site is near zero meters. In addition, to decrease the influence of terrain relief (e.g., the height of buildings), a projection plane with an altitude of negative one hundred meters is used for motion generation. The results are shown in Figure 7, where red dots and yellow dots represent starting points and terminal points of motions; and each motion is represented by a blue line linking a start point and a terminal point. The motions reveal the geometrical relationship of initial candidate matches in the object-space, which is modeled as a 2D-translation instead of the complex transformation in the image-space. By checking the motions generated from the four image pairs, we can see that: (1) the motions of some matches are approximately parallel to each other; however, for the other motions, both the direction and the length are randomly assigned. This can be verified by the motions of the first image pair in which a large number of motions have opposite directions, as presented in the top-left part of Figure 7(a); (2) the discriminative ability of the motion direction is stronger than that of the motion length; thus, the HMCC algorithm depends on a hierarchical strategy from the motion direction to the motion length for outlier elimination. This observation provides a clue to remove outliers based on analysis of motions in the object-space and forms the foundation of the HMCC algorithm for outlier elimination.

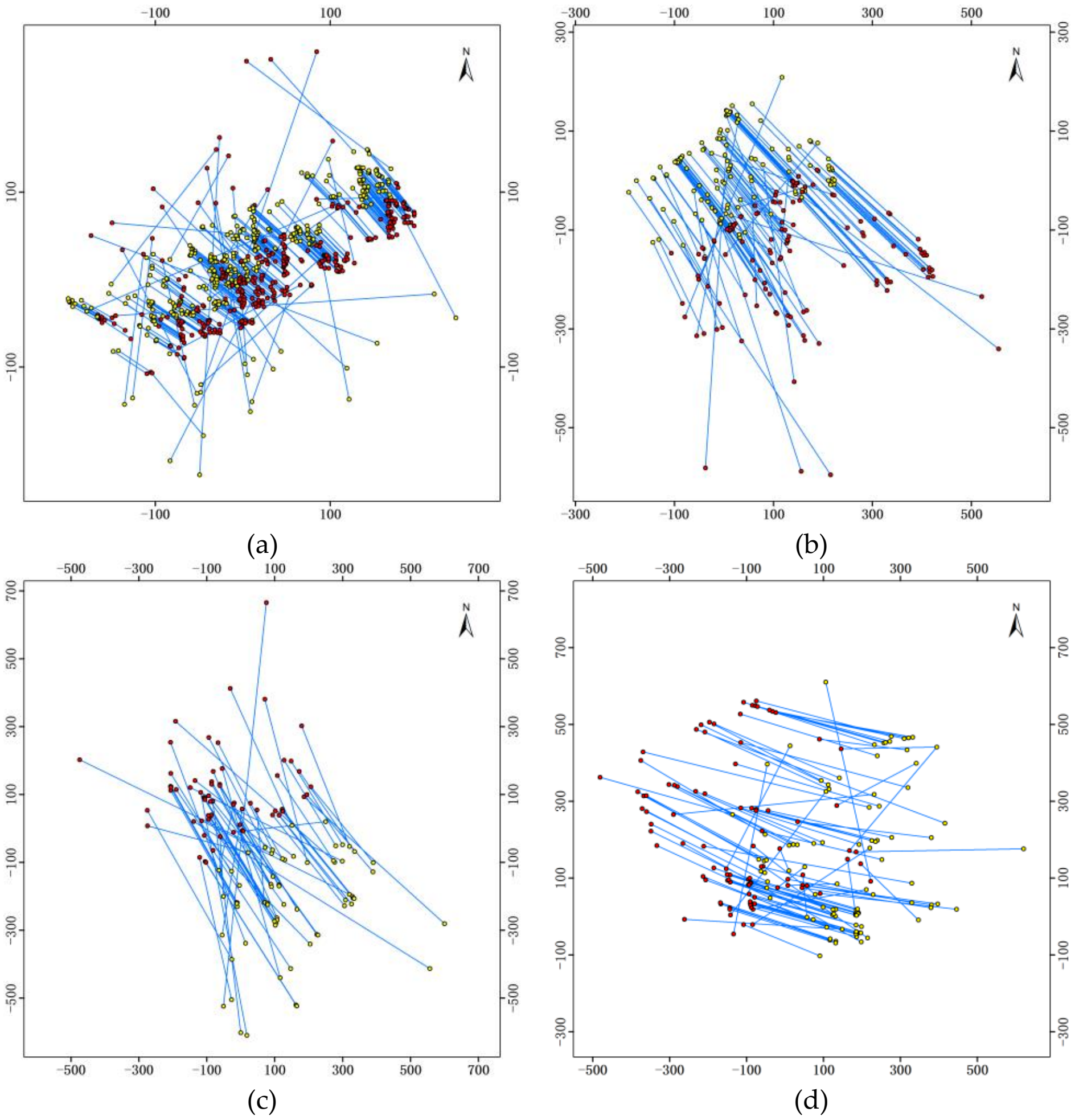

**Figure 7.** Motions generated from initial candidate matches: **(a)**, **(b)**, **(c)** and **(d)** are motions of the first, second, third and fourth image pair, respectively.

*4.3 Outlier elimination based on the HMCC algorithm*

The HMCC algorithm consists of three major steps. First, the global direction consistency constraint is used to detect motions that are markedly different in the motion direction. A large proportion of initial candidate matches would be contaminated by outliers, especially for pairs of oblique images, as shown in Figure 7(c) and Figure 7(d). Thus, the Hough voting scheme for the motion direction is utilized in the first stage due to its robustness to random noises. The results of the first-stage outlier elimination are shown in Figure 8, where motions rendered in green indicate inliers; motions rendered in red are considered outliers. The results clearly show that the red-colored motions have randomly assigned directions, which are obviously different from the motions in green; for the first and second image pairs, almost all motions with random directions can be detected, mainly because of the strong direction consistency within inliers, as illustrated in Figure 8(a) and Figure 8(b). However, due to larger perspective deformations, the direction consistency of the last two image pairs is weaker than that of the first two image pairs, which results in the partial detection of the motions with random directions, as presented in Figure 8(c) and Figure 8(d). Although the weaker consistency observed from the motions of the last two image pairs, the varying range of directions is limited, which can be deduced from the voting results of the global direction, as shown in Figure 9 (Notice that, the accumulation array is extended by copying the first nine elements to the end due to the cyclicity of the motion direction). For each image pair, the number of salient peak bins does not greater than two. Considering that the sampling interval of the voting space is set as 20°, the varying range of directions with respect to the peak bins does not exceed 40°. In addition, for the four image pairs, salient peaks can be easily determined from the accumulation arrays. In this stage, the numbers of detected outliers are 46, 21, 22 and 28 for the four image pairs, respectively, which are 12.0%, 15.7%, 32.4% and 28.9% of the corresponding total matches.

The second step of the HMCC algorithm is to remove remaining outliers by using the local direction-change constraint based on the observation that the direction-change of one motion with respect to its neighboring motions is restricted to a limited range. Similarly, to cope with the influence of outliers, a voting scheme with a strict parameter set is utilized. The results are shown in Figure 10, where motions in both green and red are generated from inliers of the global direction consistency constraint. Thus, in this stage, the estimation precision of local direction-changes can be increased by the high inliers ratios, when cooperating with the technique that the median of direction-changes is considered as the expected value of the target motion. From the results of outlier detection, we can see that: (1) only a few outliers are detected for the first and second image pairs, as shown in Figure 10(a) and Figure 10(b), because most of them have been filtered based on the direction consistency constraint. (2) For the third and fourth image pairs, most of the remaining motions with inconsistent direction-changes are detected, which can further increase inlier ratios. Similarly, the accumulation arrays of the voting scheme for the direction-change consistency constraint are presented in Figure 11. For each accumulation array, the peak bin can be noticeably observed, and almost all direction-changes of motions are limited to the first bin. We can conclude that a majority of direction-changes do not exceed 3°, because the sampling interval of the voting space in this stage is configured as 3°. Thus, the inlier ratios are dramatically increased after the use of above-mentioned voting schemes.

For the third stage, z-score tests are conducted to eliminate outliers that have almost the same directions as the other motions, but having obviously different motion lengths, which are prone to be outliers. For the four image pairs, only one motion does not pass the z-score test, which is labeled in Figure 10(b). Using the HMCC algorithm for outlier elimination, a majority of outliers can be detected from initial candidate matches, and the reduced correspondences are presented in Figure 12. For each image pair, almost all green lines are parallel to each other, which indicates that no obviously inconsistent matches exist. Therefore, the HMCC algorithm can dramatically increase inlier ratios of the four image pairs.

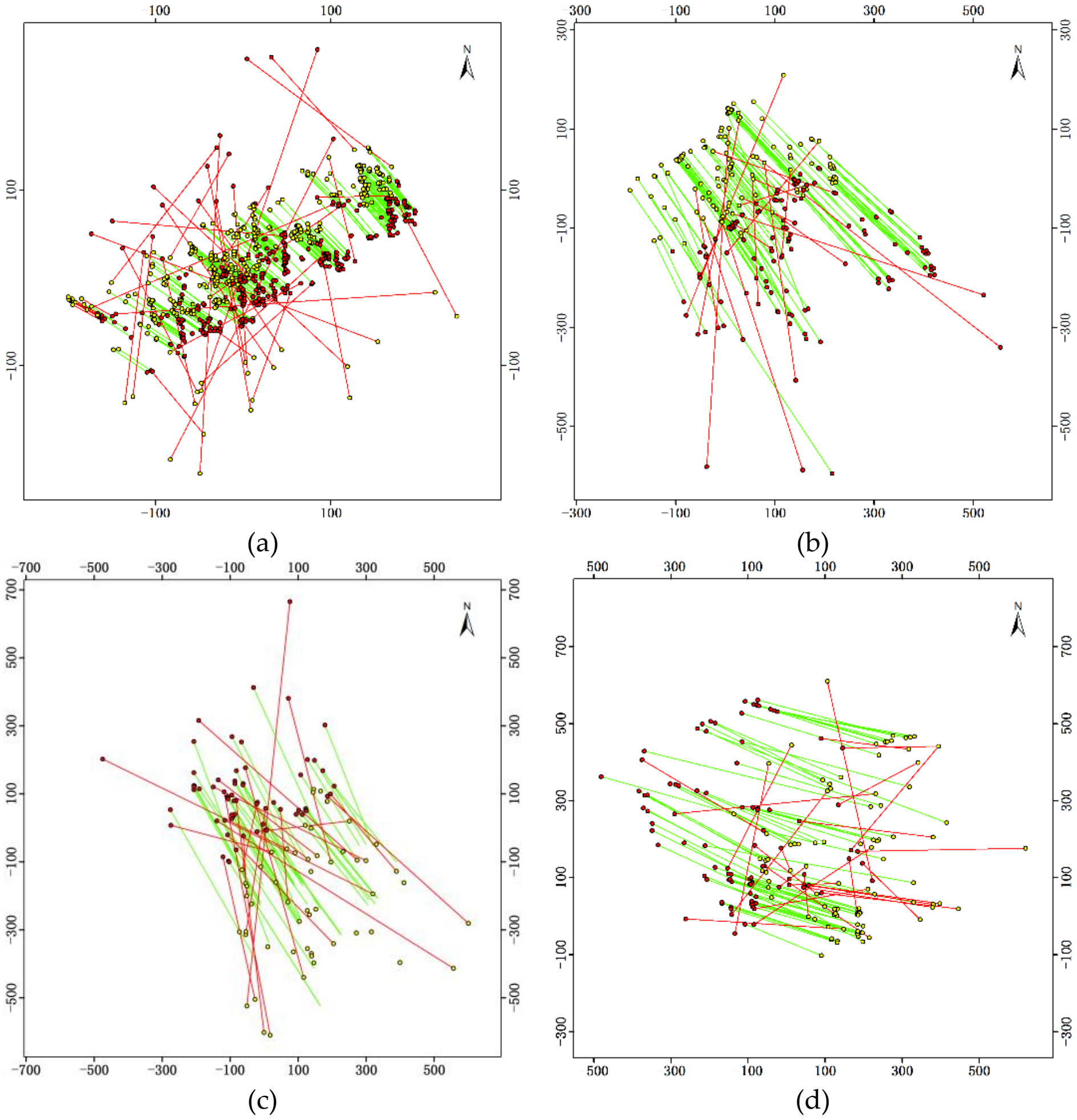

**Figure 8.** The global direction consistency constraint for the first-stage outlier elimination: **(a)**, **(b)**, **(c)** and **(d)** are results of the first, second, third and fourth image pair, respectively.

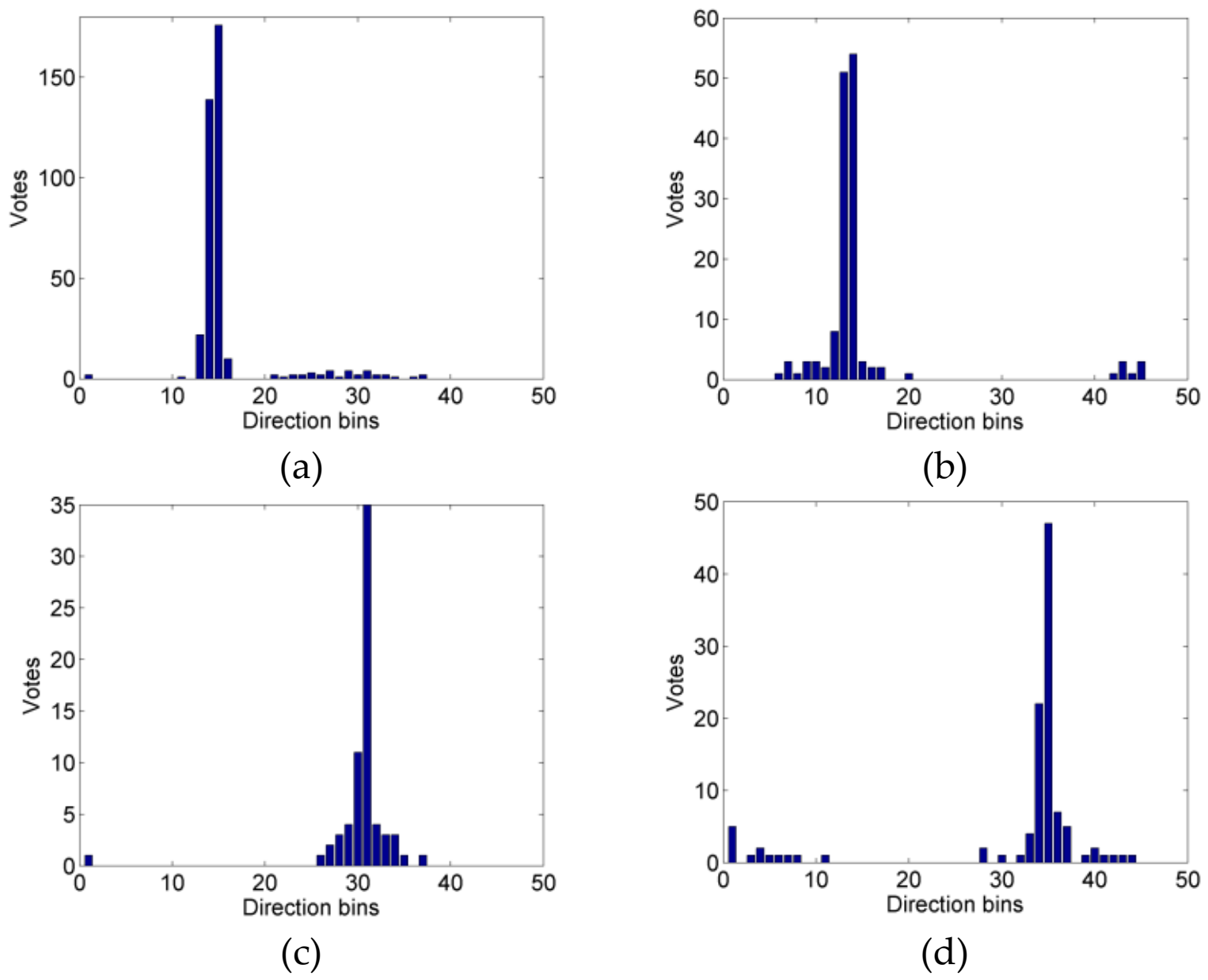

**Figure 9.** The voting results for the global direction consistency constraint: **(a)**, **(b)**, **(c)** and **(d)** correspond to the result of the first, second, third and fourth image pair, respectively.

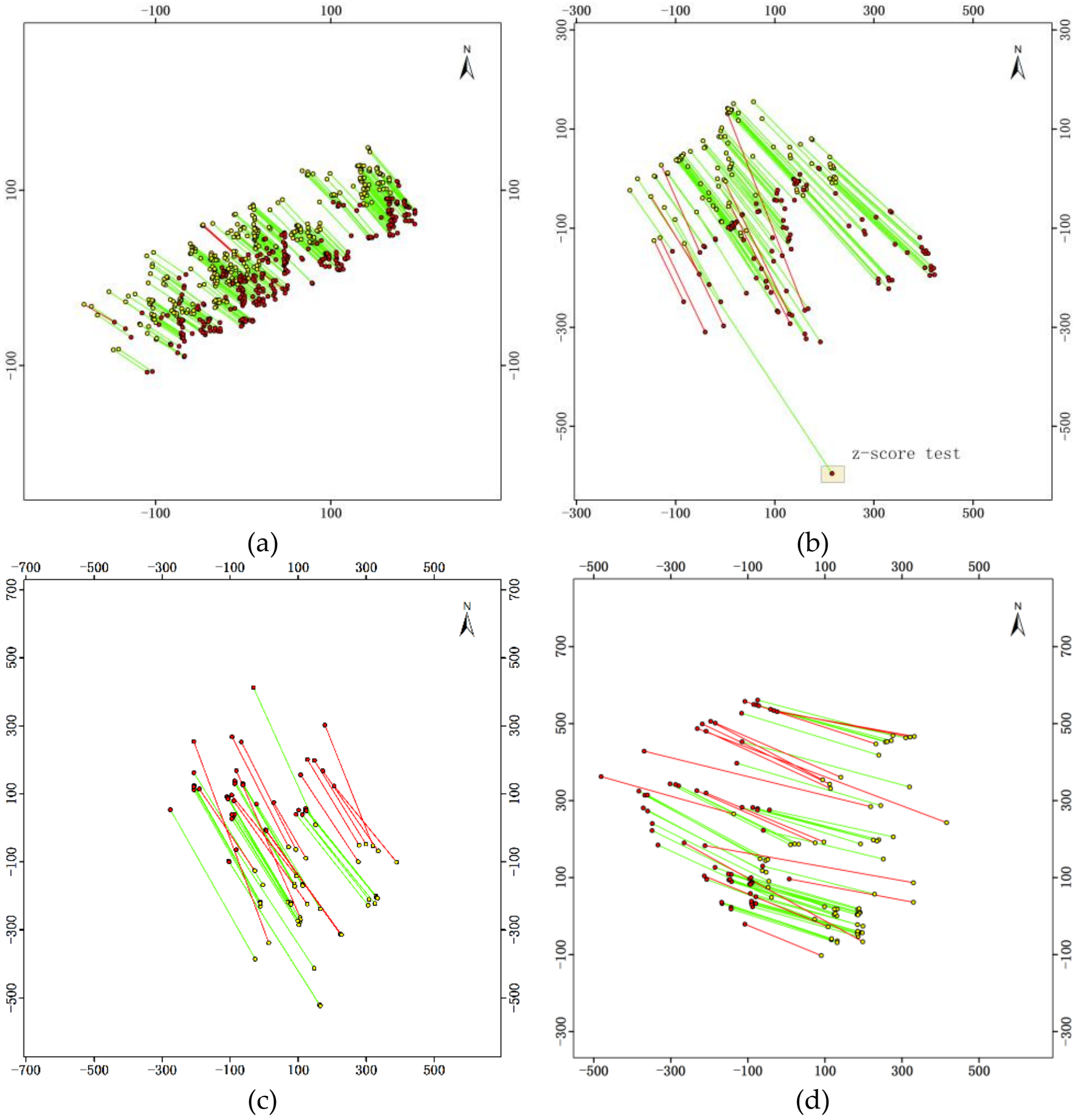

**Figure 10.** The local direction-change consistency constraint for the second-stage outlier elimination: **(a)**, **(b)**, **(c)** and **(d)** are results of the four image pairs, respectively.

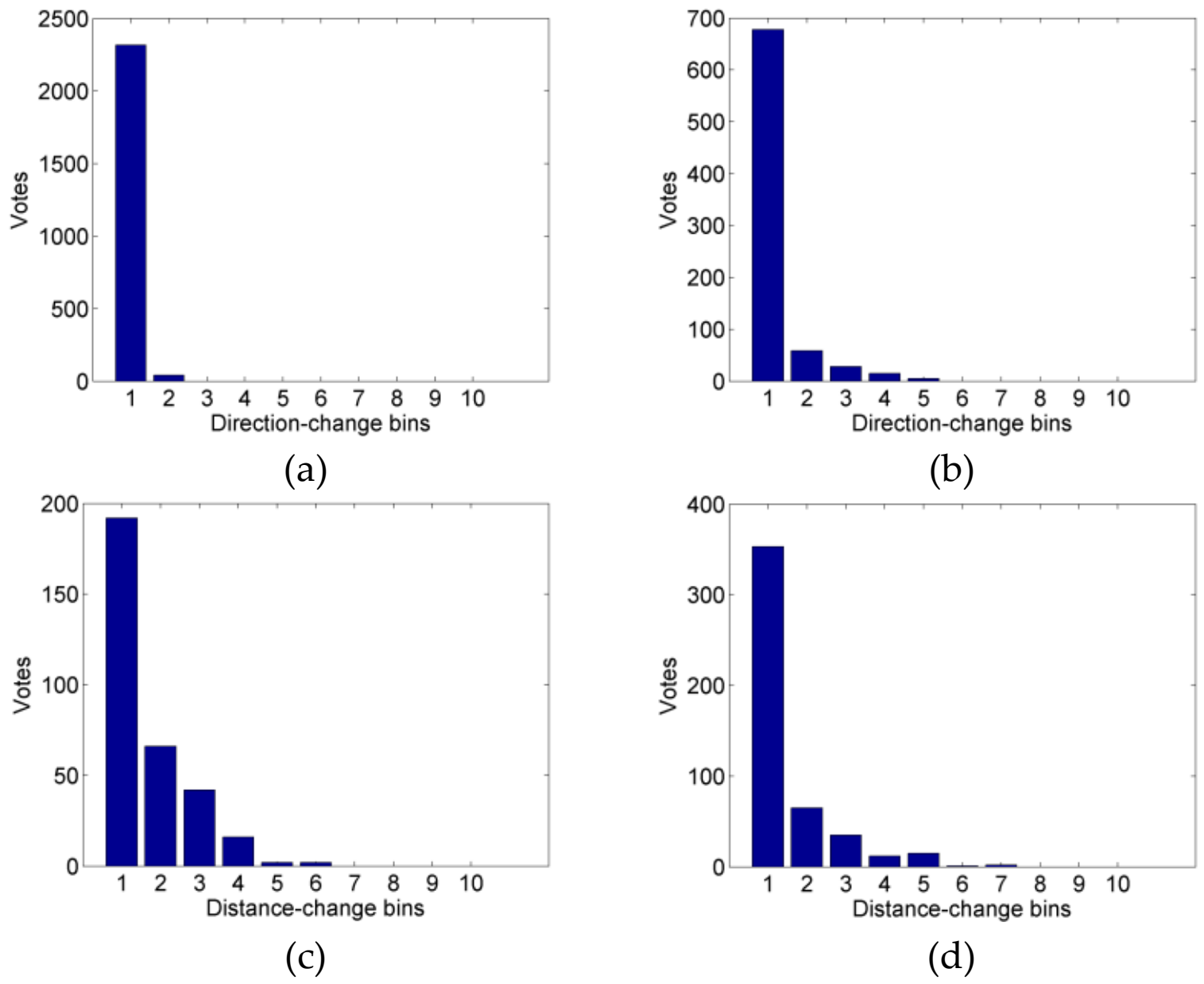

**Figure 11.** The voting results for the local direction-change consistency constraint: **(a)**, **(b)**, **(c)** and **(d)** correspond to the result of the first, second, third and fourth image pair, respectively.

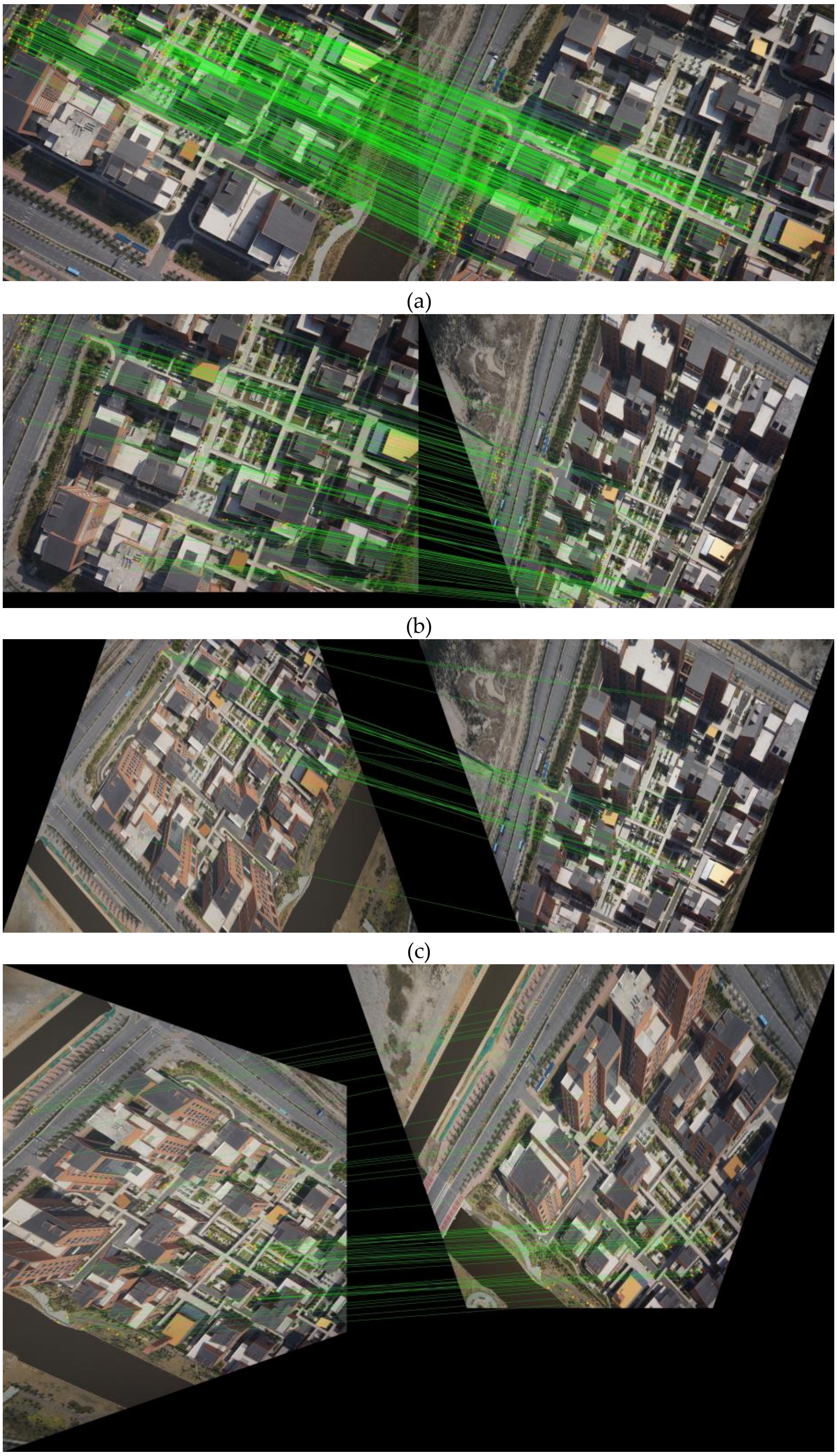

(a)

(b)

(c)

(d)

**Figure 12.** Reduced correspondence set of the four image pair: **(a)**, **(b)**, **(c)** and **(d)** are the reduced correspondence sets of the four image pairs, respectively.

*4.4. Analysis of the influence of outliers on the HMCC algorithm*

The main purpose of the HMCC algorithm is to increase the inlier ratio of initial candidate matches, which tend to be dominated by outliers, especially for oblique image pairs with large photometric and geometrical deformations. Therefore, the robustness to outliers is an essential indicator to evaluate its performance. In this section, we design a strategy to generate matches with a specified outlier ratio and analyze the algorithm's robustness to outliers.

To prepare experimental datasets, reduced correspondence sets of the four image pairs are first refined by using the RANSAC method with the estimation of a fundamental matrix, where the maximum residual error is configured as 1.0 pixel to ensure high inlier ratios. Then, outliers surviving from the epipolar constraint are further eliminated by manual inspection, which produces the total inliers for each image pair. Finally, we can generate desired matches with a specified outlier ratio by adding random point pairs to corresponding inliers. For the four image pairs, the numbers of the total inliers are 204, 44, 15 and 20, respectively. Outlier ratios ranging from 0.1 to 0.9 are evenly sampled with the interval value of 0.1.

To evaluate the robustness to outliers, the HMCC algorithm is executed on the artificially contaminated matches for the first-stage outlier elimination, and the RANSAC method with the same parameters as that for match refinement is subsequently applied to verify the results of the HMCC. Two criteria, namely, precision and recall, are used for performance evaluation. Precision is the ratio of the numbers of inliers, which are generated from the RANSAC method and the HMCC algorithm, respectively. Similarly, recall is defined as the ratio of the number of inliers generated from the RANSAC method to the number of corresponding total inliers. For the four image pairs, statistical results of precision and recall are respectively shown in Figure 13 and Figure 14. It is clearly shown that high precision can be achieved when outlier ratios are below 0.8 for the four image pairs; the precision of the first and second image pairs is better than that of the last two image pairs even with large outlier ratios reaching to 0.9, due to their relative moderate deformations caused by oblique angles. In addition, by observing the statistical results of recall, as illustrated in Figure 14, we can see that the metric recall is scarcely affected by outlier ratios, which can be deduced from the slight changes of recall. For the first and third image pairs, the mean of recall is greater than 90%; for the second image pair, the mean of recall is near 80%. However, the mean of recall is approximately 65% for the fourth image pair, because large perspective deformations cause the weak consistency of motions. This can further lead to the elimination of matches in the analysis of the local direction-changes. In conclusion, the HMCC algorithm can robustly remove a majority of outliers and noticeably increase inlier ratios when outlier ratios are not greater than 0.8; besides, this algorithm can preserve a high percentage of inliers.

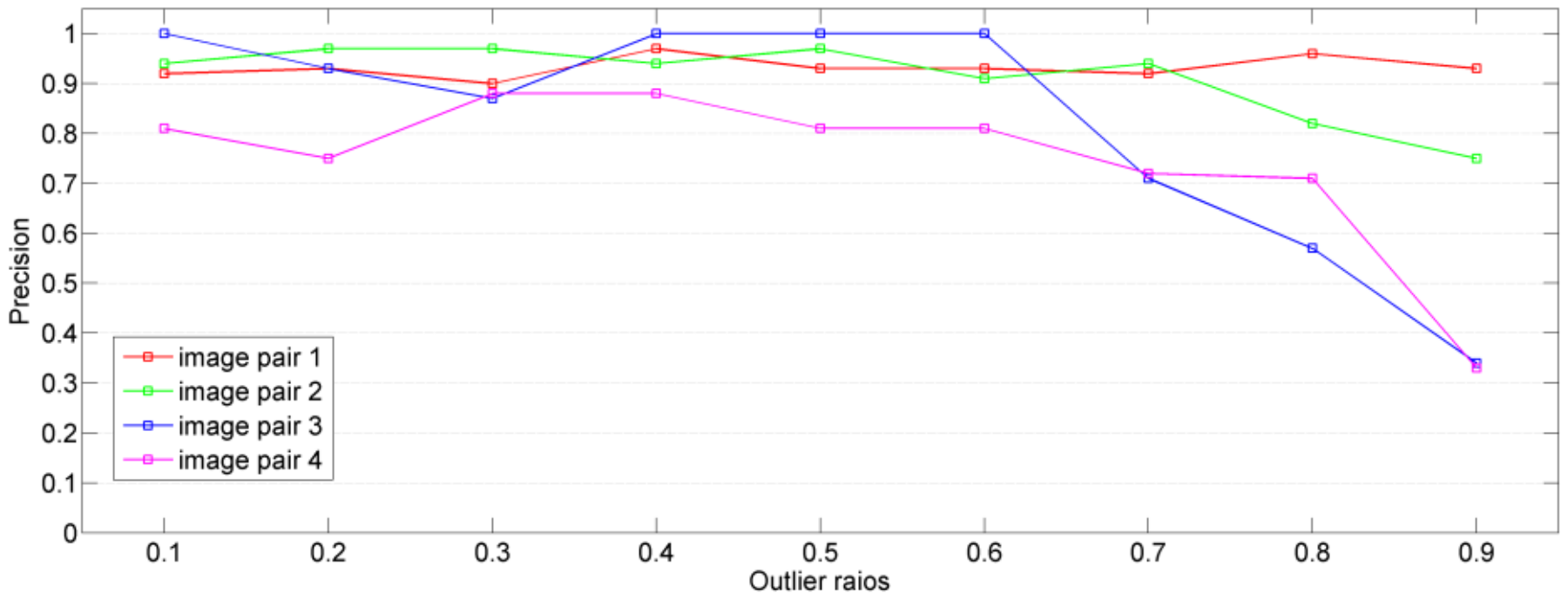

**Figure 13.** Statistical results of precision for the four image pairs.

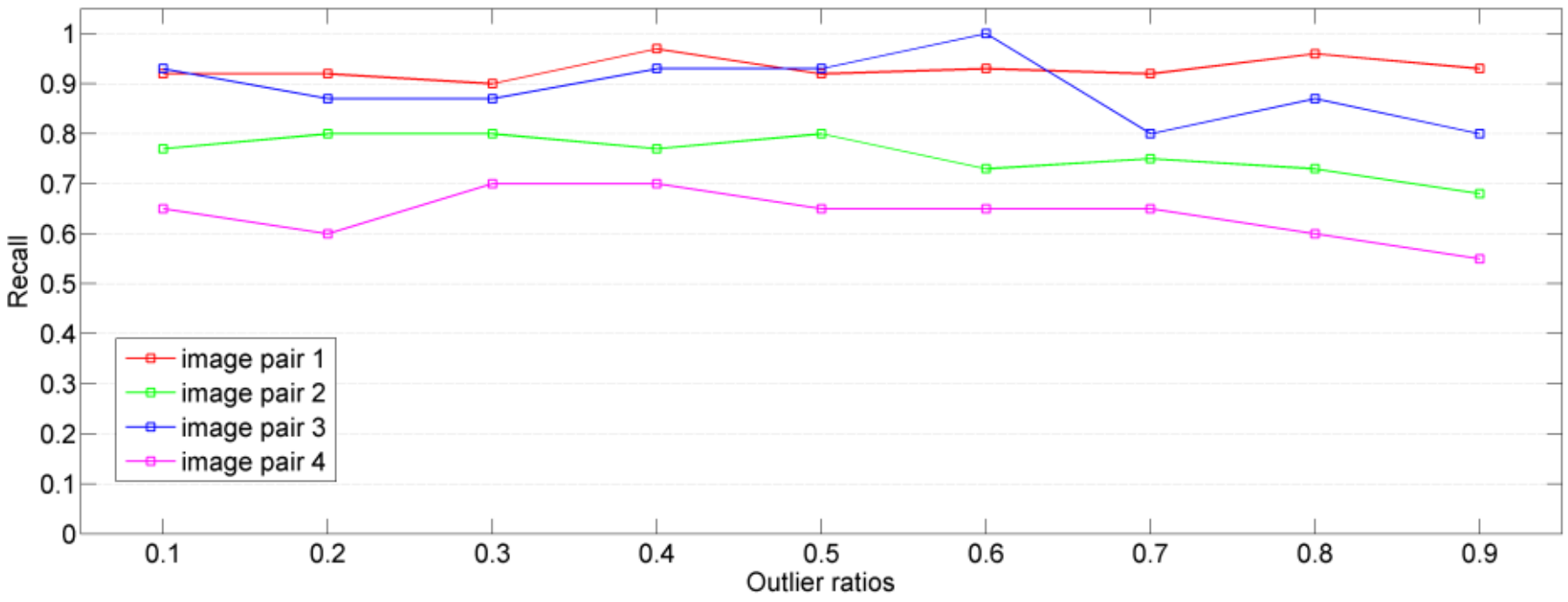

**Figure 14.** Statistical results of recall for the four image pairs.

*4.5. Comparison with various geometrical verification methods*

In this section, two methods, namely, LO-RANSAC (Chum et al., 2003) and GC-RANSAC (Lu et al., 2016) are compared with HMCC-RANSAC to assess their performance in geometrical verification. First, the four image pairs, as described in Section 4.1, are used to evaluate their performance in the filter and verification stages of three methods. Second, an image orientation test is conducted to assess their performance in terms of efficiency, completeness, and accuracy. In this study, the three methods are efficiently implemented by using the C++ programming language, and all experiments are executed on an Intel Core i7-4770 PC on the Windows platform with a 3.4 GHz CPU and a 2.0 G GeForce GTX 770M graphics card.

4.5.1. Comparison using the four image pairs

LO-RANSAC uses a local optimization stage to decrease the number of samples drawn. Similar to our method, GC-RANSAC devises a candidate match filter to increase inlier ratios based on the Hough voting scheme. Thus, performance comparison for the filter is first conducted between GC-RANSAC and HMCC-RANSAC, and results are shown in Figure 15. It is clearly shown that constant computation costs are consumed by HMCC-RANSAC; however, for GC-RANSAC, time consumption is positively proportional to the number of matches. In addition, numbers of retained matches from GC-RANSAC is greater than that from HMCC-RANSAC, mainly because of the usage of more rigorous constraints in HMCC. The statistical results of time costs are listed in Table 3.

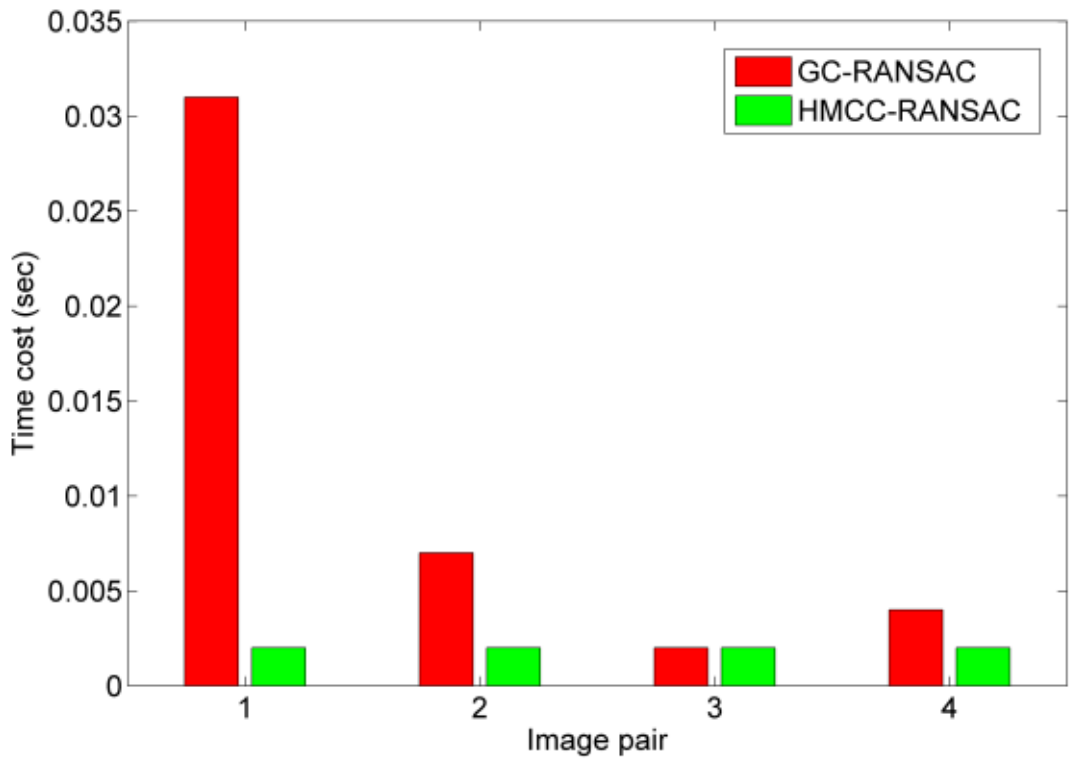

(a)

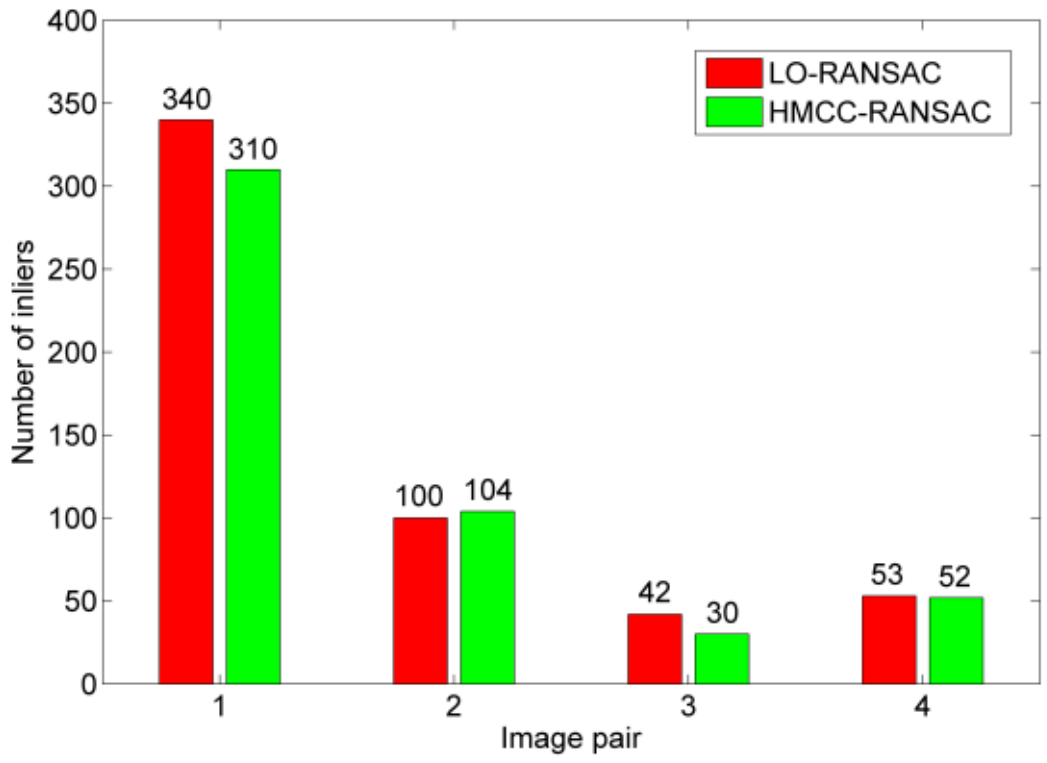

(b)

**Figure 15.** Comparison of time costs and numbers of inliers in the filter stage: **(a)** time costs, and **(b)** numbers of inliers.

Figure 16 shows the comparison results of time costs and numbers of inliers for geometrical verification with the estimation of a fundamental matrix, and the statistical results are listed in Table 3. For both GC-RANSAC and HMCC-RANSAC, the classical RANSAC method is used for the transformation estimation. The results show that the efficiency can be dramatically improved by using the filter for GC-RANSAC and HMCC-RANSAC, especially for the third and fourth image pairs with lower inlier ratios. In addition, comparative efficiency is achieved in GC-RANSAC and HMCC-RANSAC for the second image pairs; however, for the first, third and fourth image pairs, HMCC-RANSAC achieves approximately three times higher efficiency than that of GC-RANSAC, because more outliers are detected in HMCC-RANSAC based on the local direction-change consistency constraint.

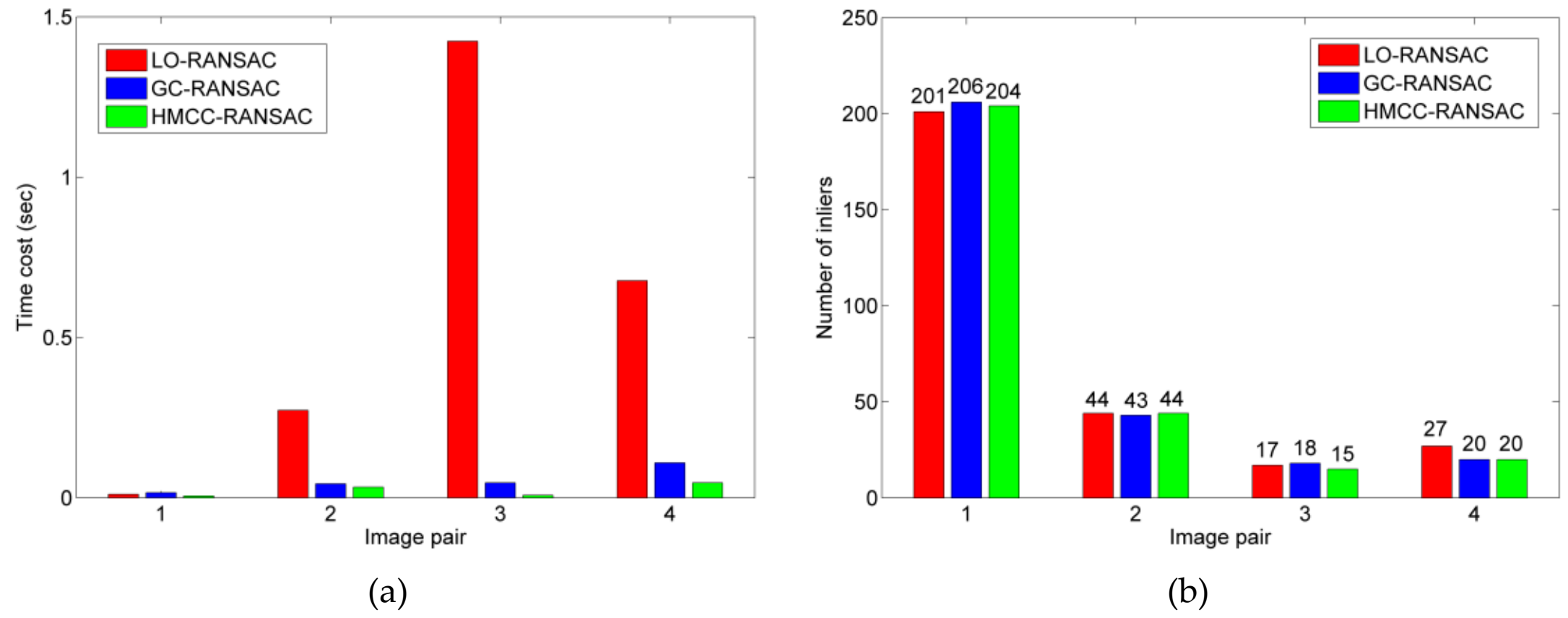

**Figure 16.** Comparison of time costs and numbers of inliers in the geometrical verification stage: **(a)** time costs, and **(b)** numbers of inliers.

**Table 3.** Statistical results of time costs for the four image pairs (unit in seconds).

| No. | LO-RANSAC | GC-RANSAC | | | HMCC-RANSAC | | |
|---|---|---|---|---|---|---|---|
| | | Filter | Verif | Sum | Filter | Verif | Sum |
| 1 | 0.011 | 0.031 | 0.016 | 0.047 | 0.002 | 0.006 | 0.008 |
| 2 | 0.273 | 0.007 | 0.044 | 0.051 | 0.002 | 0.033 | 0.035 |
| 3 | 1.424 | 0.002 | 0.047 | 0.049 | 0.002 | 0.008 | 0.010 |
| 4 | 0.677 | 0.004 | 0.109 | 0.113 | 0.002 | 0.047 | 0.049 |

#### 4.5.2. Comparison using bundle adjustment tests

To assess the performance of the three methods for UAV image matching, the comparison is also conducted in an image orientation test. Image pairs are first selected by using an overlap principle, where one image pair would be preserved only if the dimension of overlap exceeds half of the footprint size (Jiang and Jiang, 2017a). For the used dataset, a total number of 4130 pairs are selected. Then, feature matching is executed, and the comparison results of the three methods for geometrical verification is presented in Figure 17, where the time costs of the filter and verification stages for GC-RANSAC are shown in Figure 17(b) and Figure 17(c), respectively; Figure 17(d) and Figure 17(e) for HMCC-RANSAC. In addition, statistical results of time costs using four metrics, namely, Max, Average, Stddev and Sum, are listed in Table 4.

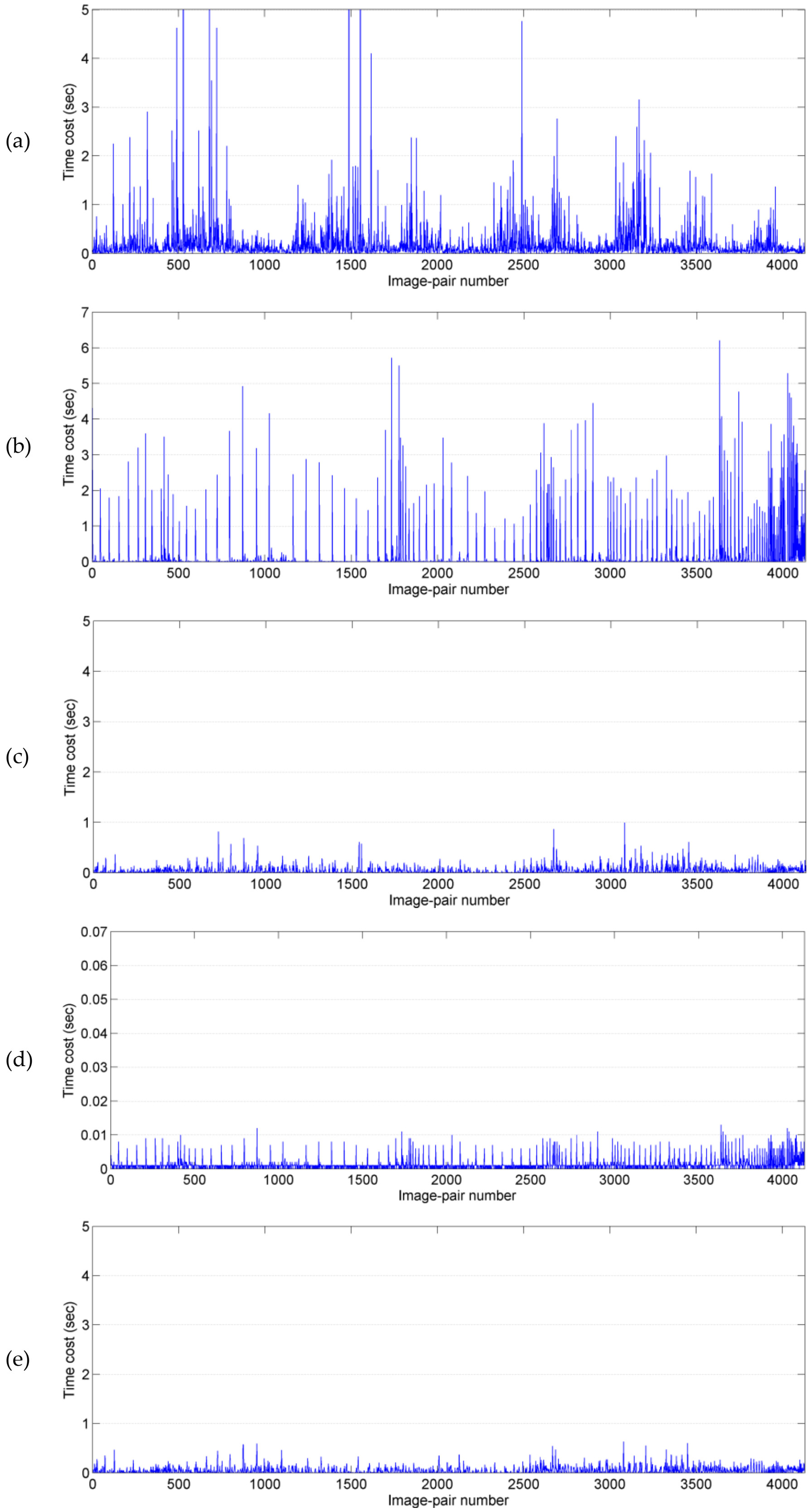

**Figure 17.** Efficiency comparison of the three methods: **(a)** time costs of LO-RANSAC; **(b)** and **(c)** time costs of GC-RANSAC in filter and verification stages; **(d)** and **(e)** time costs of HMCC-RANSAC in filter and verification stages.

It is clearly shown that when directly using LO-RANSAC, the time costs vary dramatically; for some image pairs, more than 5 seconds are consumed in the verification stage due to low inlier ratios of initial candidate matches, as illustrated in Figure 17(a). On the contrary, by using the pre-processing step, time costs of the verification with RANSAC do not exceed one second in both GC-RANSAC and HMCC-RANSAC, as shown in Figure 17(c) and Figure 17(e). By comparing the results between GC-RANSAC and HMCC-RANSAC, we can conclude that: (1) constant and low time costs are consumed in HMCC-RANSAC, as shown in Figure 17(d); however, for GC-RANSAC, high time costs can be observed from the filter step, as presented in Figure 17(b); (2) by comparing the results shown in Figure 17(c) and Figure 17(e), HMCC-RANSAC achieves higher efficiency than GC-RANSAC for some image pairs, because the local direction-change constraint can further increase inlier ratios; (3) although efficient RANSAC can be achieved by using the filter step in GC-RANSAC, the sum of time costs is not noticeably decreased because of the relatively higher time complexity in the filter stage; however, for HMCC-RANSAC, high efficiency can also be achieved in the filter stage with a speedup ratio greater than one hundred, which is verified by the mean and sum metrics in Table 4.

**Table 4.** Statistical results of efficiency comparison (unit in seconds).

| Item | LO-RANSAC | GC-RANSAC | | HMCC-RANSAC | |
|---|---|---|---|---|---|
| | | GC | RANSAC | HMCC | RANSAC |
| Max | 19.423 | 6.207 | 0.996 | 0.013 | 0.635 |
| Mean | 0.180 | 0.118 | 0.038 | 0.001 | 0.039 |
| Stddev | 0.487 | 0.495 | 0.068 | 0.001 | 0.063 |
| Sum | 743.697 | 492.537 | 155.531 | 4.395 | 162.327 |

**Table 5.** Statistical results of completeness and accuracy comparison (RMSE in pixels).

| Item | LO-RANSAC | GC-RANSAC | HMCC-RANSAC |
|---|---|---|---|
| No. points | 86,110 | 85,491 | 82,472 |
| RMSE | 0.552 | 0.542 | 0.545 |

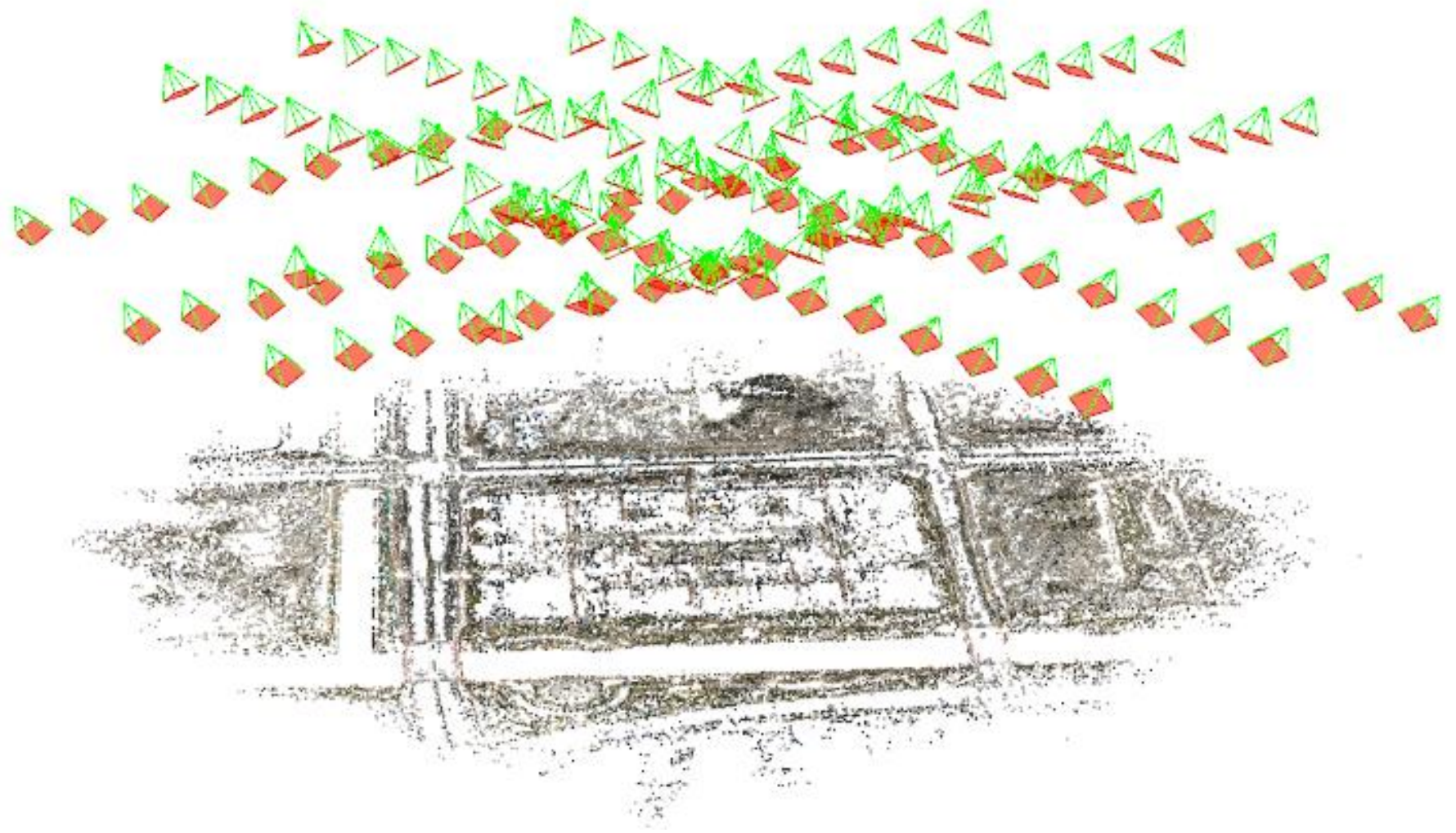

**Figure 18.** The reconstructed model by using HMCC-RANSAC for geometrical verification. The Red rectangle represents the image plane and the green line links the projection center and one of the corners of the image plane.

After feature matching of the selected image pairs, image orientation is subsequently conducted to resume accurate image poses and reconstruct 3D points of the test site. Table 5 presents the statistical results of the experiment in terms of the numbers of points and the RMSE in pixels. For lack of ground-truth data in this test site, only relative orientation is considered for performance comparison. The results indicate that image orientation can be successfully implemented with one of the three methods for geometrical verification; the RMSEs of image orientation with GC-RANSAC and HMCC-RANSAC are comparable, which is slightly better than that with LO-RANSAC. However, the largest number of points are reconstructed by LO-RANSAC, because a proportion of true matches are inevitably eliminated in the filter stage of both GC-RANSAC and HMCC-RANSAC. For the proposed algorithm in this study, the number of lost points is approximately 5 percent when compared with the number of points recovered from LO-RANSAC. In other words, with the completeness sacrifice of approximately 0.05 times the total number of points, HMCC-RANSAC can improve the verification efficiency with a speedup ratio near 5 for the dataset. The reconstructed model by using the HMCC-RANSAC for geometrical verification is shown in Figure 18.

In order to assess the transferability of the proposed algorithm, image orientation tests of additional three datasets are carried out. The three datasets cover different landscapes: the first dataset is collected from a suburban area covered by vegetation and crossed by some railroad tracks; the second test site is a farmland with dominant bare lands; the third test site locates in an urban region with a central shopping plaza surrounded by high residual buildings. For data acquisition, a multi-rotor UAV equipped with three different oblique photogrammetric systems is adopted, where an imaging system equipped one camera with 25° and -15° for pitch and roll angles is designed for data acquisition of the first dataset; an imaging system equipped two cameras for the second dataset; and a penta-view imaging system for the third dataset. The detailed information for the three campaigns is shown in Table 6. The numbers of collected images are 320, 390 and 750 for the three datasets, respectively. Based on the above-mentioned overlap principle, 5,169, 5,842 and 18,283 image pairs are selected from the first, second and third dataset, respectively. Thus, image orientation tests are performed on the datasets after feature matching of selected image pairs.

**Table 6.** Specification of image orientation tests for the additional three datasets.

| **Item** | | **Dataset 1** | **Dataset 2** | **Dataset 3** |
|---|---|---|---|---|
| *(a) Data acquisition* | | | | |
| UAV type | | multi-rotor | multi-rotor | multi-rotor |
| No. cameras | | 1 | 2 | 5 |
| Camera mount angle (°) | | front: 25, -15 | front: 25, -15<br>back: 0, -25 | nadir: 0<br>oblique: 45/-45 |
| Camera mode | | Sony RX1R | Sony RX1R | Sony NEX-7 |
| Image size (pixel×pixel) | | 6000×4000 | 6000×4000 | 6000×4000 |
| No. images | | 320 | 390 | 750 |
| *(b) Image orientation* | | | | |
| LO-RANSAC | Time (sec) | 126.0 | 198.5 | 2604.2 |
| | No. points | 173,670 | 297,884 | 309,867 |
| | RMSE (pixel) | 0.601 | 0.513 | 0.659 |
| GC-RANSAC | Time (sec) | 1445.0 (1336.4/108.5) | 2359.4 (2248.7/110.7) | 1247.2 (990.6/256.6) |
| | No. points | 171,092 | 294,639 | 300,144 |
| | RMSE (pixel) | 0.597 | 0.510 | 0.655 |
| HMCC-RANSAC | Time (sec) | 128.1 (9.4/118.7) | 120.1 (13.9/106.2) | 437.2 (17.0/420.2) |
| | No. points | 162,437 | 280,952 | 283,483 |
| | RMSE (pixel) | 0.592 | 0.500 | 0.654 |

The statistical results of image orientation are listed in Table 6, and three metrics, including the time involving in geometrical verification, the number of reconstructed points and the RMSE estimated from bundle adjustment, are used for transferability assessment. Similarly, the three methods are compared in the image orientation tests of the three datasets. Notice, for GC-RANSAC and HMCC-RANSAC, the metric time consist of three parts, namely the total time and the time consumed respectively in filter and verification stages (values in the round bracket). We can conclude that: (1) for the three datasets, HMCC-RANSAC can achieve high efficiency for geometrical verification with lower time consumption of 128.1 sec, 120.1 sec and 437.2 sec for the first, second and third dataset, respectively, and the average time costs do not exceed 0.002 sec in the filter stage; (2) comparable time costs between LO-RANSAC and HMCC-RANSAC can be observed from the first two datasets with a speedup ratio less than 2.0; however, for the third dataset, the ratio of time costs is approximately 6.0, mainly because larger oblique imaging angles configured in the third dataset cause a majority of false matches and decrease the inlier ratio of initial candidate matches; (3) dominant time costs are consumed in the filter stage of GC-RANSAC, which can be deduced from the three tests, especially for the first and second datasets. The main reason is that many more initial matches are searched from image pairs with smaller oblique angles and consequently increase the time costs in the filter stage of GC-RANSAC; (4) although approximately 7 percent of points are lost in reconstruction models, identical orientation accuracy can be observed by using HMCC-RANSAC. The time costs for the three datasets are illustrated in Figure A1, Figure A2 and Figure A3, respectively. In conclusion, the propose algorithm in this paper, namely HMCC-RANSAC, can achieve efficient geometrical verification in UAV image matching with comparable accuracy.

## 5. Discussion

This paper proposes the HMCC-RANSAC algorithm for efficient geometrical verification in UAV image matching. By using on-board GNSS/IMU data and camera mounting angles, the complex transformation model in the image-space can be simplified as a simple 2D-translation in the object-space, which is achieved by the projection of feature points. The experimental results from real datasets demonstrate that the time consumption of geometrical verification is obviously decreased for outlier dominated matches. Compared with existing strategies, such as LO-RANSAC (Chum et al., 2003) and GC-RANSAC (Lu et al., 2016), the proposed algorithm has the following advantages.

First, the HMCC-RANSAC algorithm converts the transformation model, e.g., a similarity or affine transformation, between correspondences to a simple translation model. In the literature, some other strategies have either depended on priors, including the orientation and scale, from extracted features (Schönberger et al., 2016) or utilized the pairwise geometrical relationships for parameter estimation (Li et al., 2015; Lu et al., 2016). The former requires that special feature detectors that extract related parameters are adopted in the stage of feature extraction; the latter could consume a large proportion of time costs in the verification, which can be deduced from the efficiency analysis in Section 4.5. By the further exploitation of on-board GNSS/IMU data, the complex transformation model in the image-space is simplified, and a straightforward model, termed motion in this study, is defined and used to establish the relationship between correspondence points. Second, based on the simplified model, a motion consistent constraint is implemented in terms of motion direction and motion length, which consists of a global direction consistency constraint, a local direction-change consistency constraint and a global length consistency constraint. In contrast to a multi-dimensional voting strategy, the hierarchical motion consistency constraint (HMCC) is implemented in this study by sequentially conducting the three constraints for outlier elimination, which is achieved by one-dimensional voting. As verified in Section 4.3, the HMCC can detect a large proportion of outliers with high precision and recall. Finally, the HMCC-RANSAC algorithm does not rely on other data sources except for the on-board GNSS/IMU data from flight control systems and camera mounting angles of oblique imaging systems. These data sources can be easily accessed. Consequently, the proposed algorithm can be used with wide range of application in the field of UAV photogrammetry.

Through the efficiency analysis presented in Section 4.5.2, it is clearly shown that the proposed algorithm can accelerate geometrical verification with a speedup ratio reaching to 6 for oblique datasets; however, for datasets collected from bare land regions with relative small oblique imaging angles, the HMCC-RANSAC algorithm cannot dramatically increase the verification efficiency, because of high inlier ratios of initial candidate matches. Compared with the pairwise geometrical verification, very high efficiency can also be noticed from datasets captured by imaging systems with small oblique angles, because a computation-efficient motion model is adopted in the HMCC-RANSAC algorithm. Thus, the proposed algorithm in this paper is applicable for both nadir and oblique images.

Currently, all the used datasets for performance evaluation are collected from sites with moderate terrain relief. Therefore, projection planes are approximated by average altitudes of test sites in motion generation. To expand its application for scenarios with dramatic mountain terrain, auxiliary data sources, such as SRTM (Shuttle Radar Topography Mission) (Rodriguez et al., 2006) can be used for accurate approximations of projection planes with a height precision of 10 m in most regions. Based on the experimental results as presented in Section 4.3, it would be interesting to only utilized the HMCC algorithm for geometrical verification without the use of RANSAC method, because almost all obvious outliers are eliminated. In addition, combined with the work documented in (Jiang and Jiang, 2017a, b), a complete workflow from match pair selection to geometrical verification can be established to achieve efficient UAV image matching and orientation.

## 6. Conclusions

In this paper, we propose the HMCC-RANSAC algorithm to achieve efficient geometrical verification in UAV image matching. Feature points of initial matches are first projected onto a plane and spatial relationships between correspondence set are simplified as 2D-translation, which is modeled as motions with two attributions, namely the motion direction and length. Then, a hierarchical motion consistency constraint (HMCC) is designed for outlier elimination, which is efficiently implemented by using a two-stage voting scheme. Finally, the proposed algorithm is evaluated by comprehensive comparison and analysis from the aspects of feature matching and image orientation. As shown in the experiments, strong separation ability can be noticeably observed from motions generated from initial candidate matches, and a majority of outliers can be efficiently eliminated by using the HMCC before the consequent execution of the basic RANSAC algorithm. For UAV image matching, the algorithm proposed in this paper can achieve high efficiency in geometrical verification.

## Acknowledgment

The authors would like to thank authors who have made their algorithms of SiftGPU as a free and open-source software package, which is really helpful to the research in this paper. Meanwhile, heartfelt thanks to the anonymous reviewers and the editors, whose comments and advice improve the quality of the work.

## Appendix A

See Figure A1, Figure A1, Figure A3.

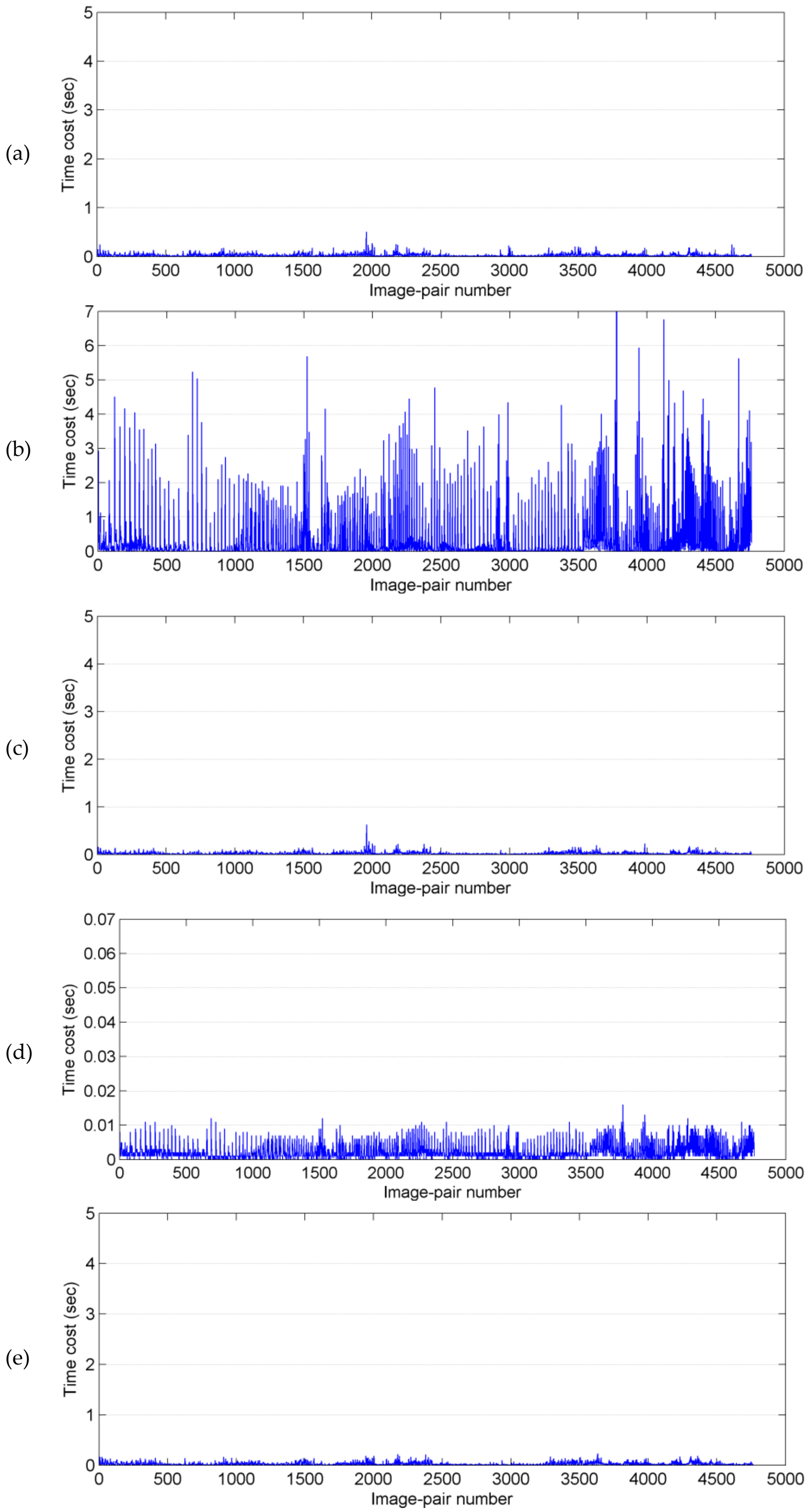

**Figure A1.** Efficiency comparison of the dataset 1: **(a)** time costs of LO-RANSAC; **(b)** and **(c)** time costs of GC-RANSAC in filter and verification stages; **(d)** and **(e)** time costs of HMCC-RANSAC in filter and verification stages.

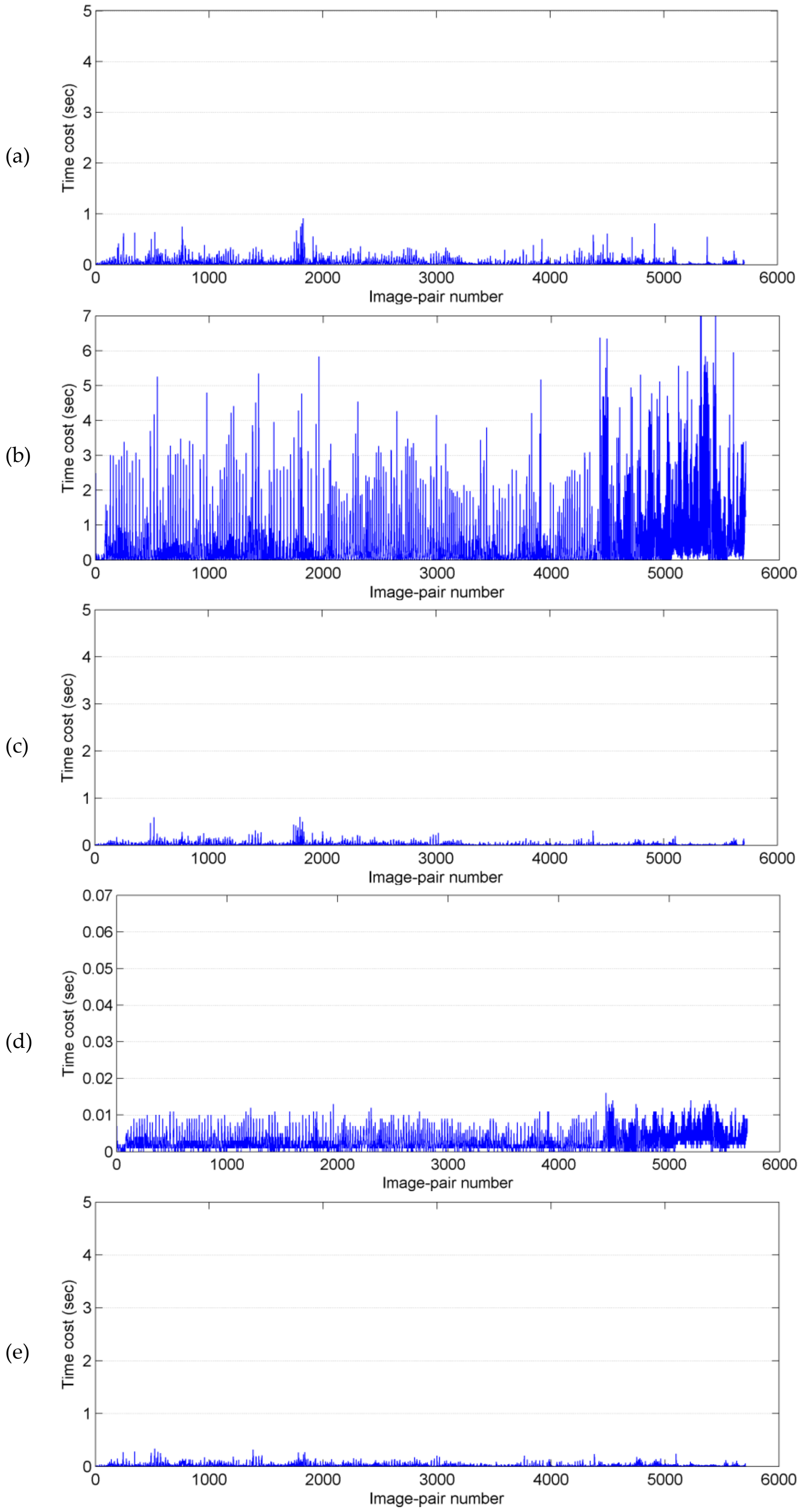

**Figure A2.** Efficiency comparison of the dataset 2: **(a)** time costs of LO-RANSAC; **(b)** and **(c)** time costs of GC-RANSAC in filter and verification stages; **(d)** and **(e)** time costs of HMCC-RANSAC in filter and verification stages.

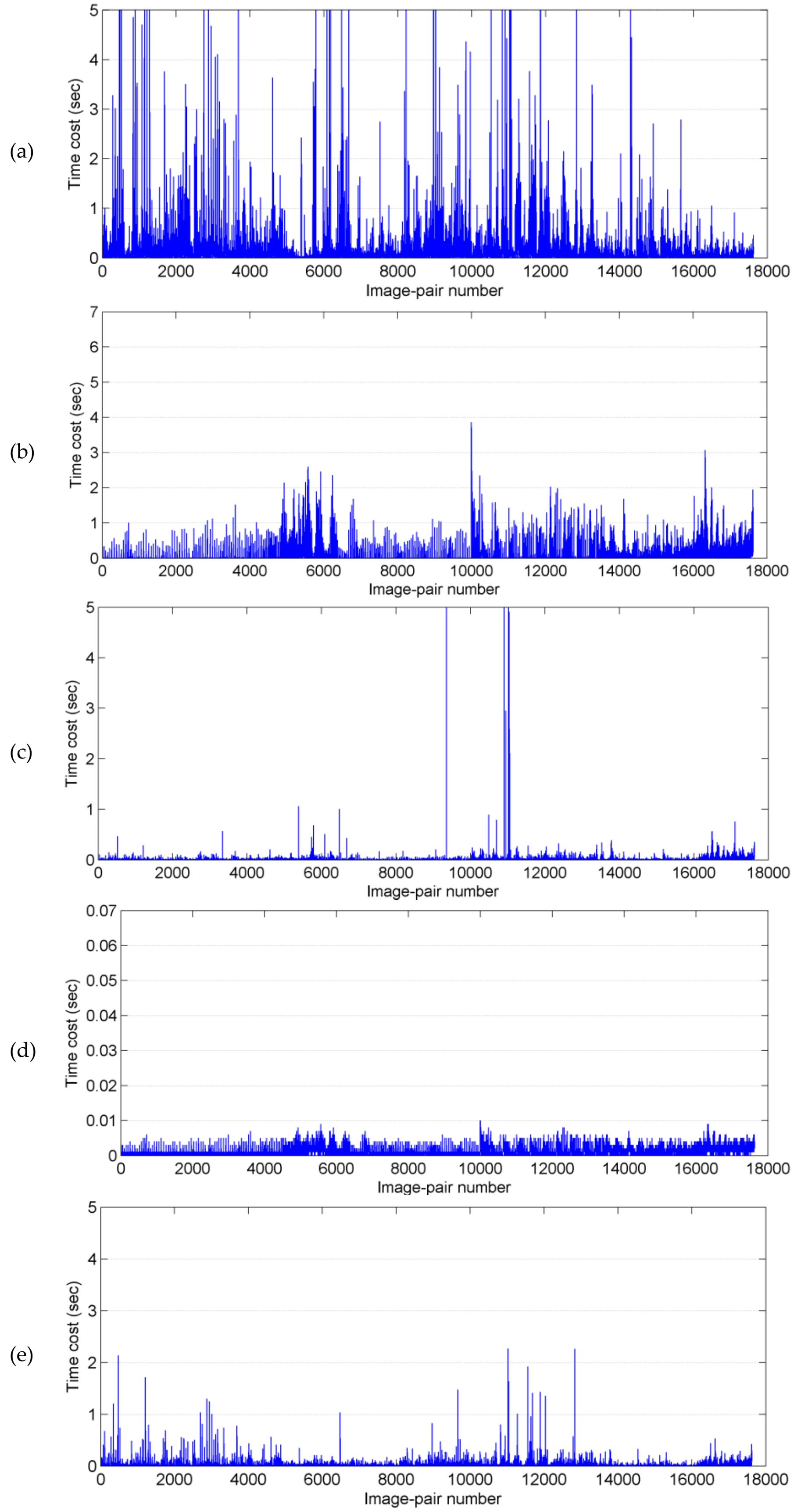

**Figure A3.** Efficiency comparison of the dataset 3: **(a)** time costs of LO-RANSAC; **(b)** and **(c)** time costs of GC-RANSAC in filter and verification stages; **(d)** and **(e)** time costs of HMCC-RANSAC in filter and verification stages.

## References

Aicardi, I., Chiabrando, F., Grasso, N., Lingua, A.M., Noardo, F., Spanò, A., 2016. UAV photogrammetry with oblique images: First analysis on data acquisition and processing. International Archives of the Photogrammetry, Remote Sensing and Spatial Information Sciences 41, 835-842.

Chum, O., Matas, J., 2005. Matching with PROSAC-progressive sample consensus, Computer Vision and Pattern Recognition, 2005. CVPR 2005. IEEE Computer Society Conference on. IEEE, pp. 220-226.

Chum, O., Matas, J., 2008. Optimal randomized RANSAC. IEEE Transactions on Pattern Analysis and Machine Intelligence 30, 1472-1482.

Chum, O., Matas, J., Kittler, J., 2003. Locally optimized RANSAC. Pattern recognition, 236-243.

Cover, T., Hart, P., 1967. Nearest neighbor pattern classification. IEEE Transactions on Information Theory 13, 21-27.

Fischler, M.A., Bolles, R.C., 1981. Random sample consensus: a paradigm for model fitting with applications to image analysis and automated cartography. Communications of the ACM 24, 381-395.

Habib, A., Han, Y., Xiong, W., He, F., Zhang, Z., Crawford, M., 2016. Automated Ortho-Rectification of UAV-Based Hyperspectral Data over an Agricultural Field Using Frame RGB Imagery. Remote Sensing 8, 796.

Harris, C., Stephens, M., 1988. A combined corner and edge detector, Alvey vision conference. Manchester, UK, p. 50.

Hough, P.V., 1962. Method and means for recognizing complex patterns. Google Patents.

Hu, H., Zhu, Q., Du, Z., Zhang, Y., Ding, Y., 2015. Reliable Spatial Relationship Constrained Feature Point Matching of Oblique Aerial Images. Photogrammetric Engineering & Remote Sensing 81, 49-58.

Jiang, S., Jiang, W., 2017a. Efficient structure from motion for oblique UAV images based on maximal spanning tree expansion. ISPRS Journal of Photogrammetry and Remote Sensing 132, 140-161.

Jiang, S., Jiang, W., 2017b. On-Board GNSS/IMU Assisted Feature Extraction and Matching for Oblique UAV Images. Remote Sensing 9, 813.

Jiang, S., Jiang, W., Huang, W., Yang, L., 2017. UAV-Based Oblique Photogrammetry for Outdoor Data Acquisition and Offsite Visual Inspection of Transmission Line. Remote Sensing 9, 278.

Li, X., Larson, M., Hanjalic, A., 2015. Pairwise geometric matching for large-scale object retrieval, Proceedings of the IEEE Conference on Computer Vision and Pattern Recognition, pp. 5153-5161.

Lowe, D.G., 2004. Distinctive image features from scale-invariant keypoints. International journal of computer vision 60, 91-110.

Lu, L., Zhang, Y., Tao, P., 2016. Geometrical Consistency Voting Strategy for Outlier Detection in Image Matching. Photogrammetric Engineering & Remote Sensing 82, 559-570.

Mikolajczyk, K., Schmid, C., 2005. A performance evaluation of local descriptors. Pattern Analysis and Machine Intelligence, IEEE Transactions on 27, 1615-1630.

Raguram, R., Chum, O., Pollefeys, M., Matas, J., Frahm, J.-M., 2013. USAC: a universal framework for random sample consensus. IEEE transactions on pattern analysis and machine intelligence 35, 2022-2038.
Raguram, R., Frahm, J.-M., Pollefeys, M., 2008. A comparative analysis of RANSAC techniques leading to adaptive real-time random sample consensus. Computer Vision–ECCV 2008, 500-513.
Rodriguez, E., Morris, C.S., Belz, J.E., 2006. A global assessment of the SRTM performance. Photogrammetric Engineering & Remote Sensing 72, 249-260.
Sattler, T., Leibe, B., Kobbelt, L., 2009. SCRAMSAC: Improving RANSAC's efficiency with a spatial consistency filter, Computer vision, 2009 ieee 12th international conference on. IEEE, pp. 2090-2097.
Schönberger, J.L., Price, T., Sattler, T., Frahm, J.-M., Pollefeys, M., 2016. A vote-and-verify strategy for fast spatial verification in image retrieval, Asian Conference on Computer Vision. Springer, pp. 321-337.
Sun, Y., Sun, H., Yan, L., Fan, S., Chen, R., 2016. RBA: Reduced Bundle Adjustment for oblique aerial photogrammetry. ISPRS Journal of Photogrammetry and Remote Sensing 121, 128-142.
Tsai, C.-H., Lin, Y.-C., 2017. An accelerated image matching technique for UAV orthoimage registration. ISPRS Journal of Photogrammetry and Remote Sensing 128, 130-145.
Turner, D., Lucieer, A., Wallace, L., 2014. Direct georeferencing of ultrahigh-resolution UAV imagery. Ieee T Geosci Remote 52, 2738-2745.
Yao, J., Cham, W.-K., 2007. Robust multi-view feature matching from multiple unordered views. Pattern Recognition 40, 3081-3099.
Zhang, Z., 1998. Determining the epipolar geometry and its uncertainty: A review. International journal of computer vision 27, 161-195.
Zhuo, X., Koch, T., Kurz, F., Fraundorfer, F., Reinartz, P., 2017. Automatic UAV Image Geo-Registration by Matching UAV Images to Georeferenced Image Data. Remote Sensing 9, 376.